\documentclass[11pt]{article}

\pdfoutput=1

\usepackage[final]{acl}

\usepackage{microtype}
\usepackage{graphicx}
\usepackage{subcaption}
\usepackage{booktabs} 
\usepackage{enumitem}

\usepackage{arydshln} 

\usepackage[most]{tcolorbox}
\usepackage{listings}
\usepackage{xcolor}
\definecolor{mylightblue}{RGB}{173,216,230} 
\usepackage{amsmath}
\usepackage{amssymb}
\usepackage{mathtools}
\usepackage{amsthm}
\usepackage{multirow}
\usepackage[utf8]{inputenc}
\usepackage{xspace}
\usepackage{bbm}
\usepackage{float}

\usepackage[textsize=tiny]{todonotes}

\newcommand{\squishlist}{
	\begin{list}{$\bullet$}
		{ \setlength{\itemsep}{1pt}
			\setlength{\parsep}{1pt}
			\setlength{\topsep}{2.5pt}
			\setlength{\partopsep}{0.5pt}
			\setlength{\leftmargin}{1em}
			\setlength{\labelwidth}{1em}
			\setlength{\labelsep}{0.6em}
		}
	}
	\newcommand{\squishend}{
	\end{list}
}

\usepackage{enumitem}

\usepackage{amsmath}
\usepackage{amsfonts}
\usepackage{algorithm} 
\usepackage{algorithmic}  
\usepackage[algo2e]{algorithm2e} 
\usepackage{pgfplots}
\pgfplotsset{compat=1.18}
\usepackage{tikz}
\usepackage{times}
\usepackage{latexsym}

\usepackage[T1]{fontenc}
\usepackage[utf8]{inputenc}
\usepackage{microtype}

\usepackage{inconsolata}

\usepackage{graphicx}

\title{RecMem: Recurrence-based Memory Consolidation for Efficient and Effective Long-Running LLM Agents}

\author{
  \textbf{Zijie Dai}$^{1}$ \quad
  \textbf{Shiyuan Deng}$^{2}$\thanks{\ Dr. Shiyuan Deng is the corresponding author.} \quad
  \textbf{Sheng Guan}$^{3}$ \quad
  \textbf{Yizhou Tian}$^{1}$ \\
  \textbf{Xin Yao}$^{4}$ \quad
  \textbf{Xiao Yan}$^{5}$ \quad
  \textbf{James Cheng}$^{1}$ \\[0.3em]
  $^{1}$Department of Computer Science and Engineering, The Chinese University of Hong Kong \\
  $^{3}$School of Computer Science, Beijing University of Posts and Telecommunications \\
  $^{2}$Huawei Cloud, \quad
  $^{4}$Huawei Theory Lab, \quad
  $^{5}$Institute for Math and AI, Wuhan University \\[0.2em]
  \texttt{caiusdai@link.cuhk.edu.hk} \quad
  \texttt{dengshiyuan@huawei.com}
}

\begin{document}
\maketitle

\begin{abstract}

Memory systems often organize user-agent interactions as retrievable external memory and are crucial for long-running agents by overcoming the limited context windows of LLMs. However, existing memory systems invoke LLMs to process every incoming interaction for memory extraction, and such an \textit{eager memory consolidation} scheme leads to substantial token consumption. To tackle this problem, we propose \textit{RecMem} by rethinking when memory consolidation should be conducted. RecMem stores incoming interactions in a subconscious memory layer and encode them using lightweight embedding models for retrieval. LLMs are only invoked to extract episodic and semantic memory when sustained recurrence are observed for semantically similar interactions. Such \textit{recurrence-based consolidation} works because these interactions correspond to a semantic cluster with rich information and thus are worth extraction and summarization. To improve accuracy, RecMem also incorporates a semantic refinement mechanism that recovers the fine-grained facts omitted by memory extraction. Experiments show that RecMem reduces the memory construction token cost of three SOTA memory systems by up to 87\% while exceeding their accuracy. Our code is available at \url{https://github.com/CaiusDai/RecMem}.

\end{abstract}

\section{Introduction}
\label{sec:introduction}

Large Language Models (LLMs) have demonstrated strong capabilities across a wide range of tasks~\cite{deepseek_code,deepseek_math}. However, enabling LLMs to function as long-running agents requires accumulating experience over extended user-agent interactions~\cite{self_evolve}. In practice, this is hindered by two critical limitations: current LLMs cannot retain information beyond their limited context windows~\cite{forget}, and they often under-utilize relevant evidences even if they are present in long inputs due to the lost-in-the-middle effect~\cite{lostinthemiddle}.


To address these limitations, memory systems emerge as an essential component for building long-running LLM agents~\cite{self_evolve,memory_reason}, and many solutions have been proposed with different memory structures and memory extraction methods~\cite{a-mem,mem0,treemem,memgpt,locomo}. For example, Zep~\cite{zep} constructs temporal knowledge graphs by abstracting relational triplets from interactions; Mem0~\cite{mem0} extracts atomic facts from interactions for similarity-based retrieval; A-Mem~\cite{a-mem} organizes interactions as connected notes, and a note can update the contents of its neighbors.

\begin{figure*}[!t]
    \centering
    \begin{subfigure}[b]{0.29\linewidth}
        \vspace{0pt}
        \includegraphics[width=\linewidth]{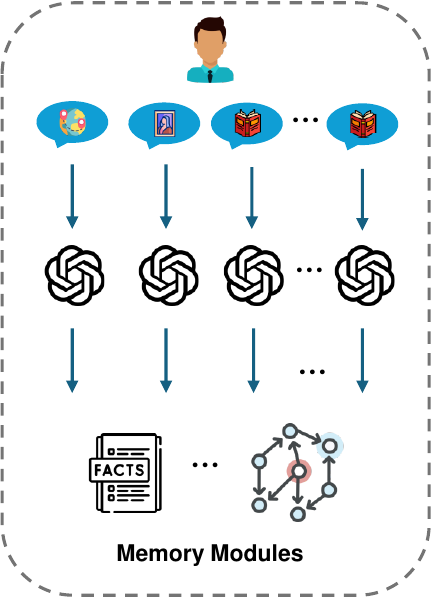}
        \caption{Eager memory consolidation}
    \end{subfigure}
    \hfill
    \begin{subfigure}[b]{0.37\linewidth}
        \vspace{0pt} 
        \includegraphics[width=\linewidth]{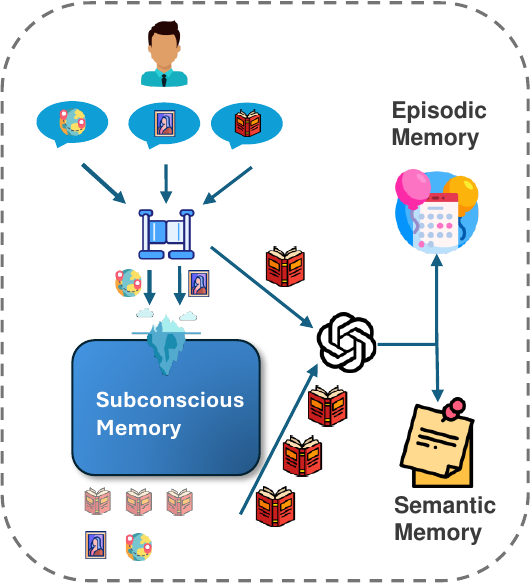}
        \caption{Recurrence-based consolidation (ours)}
    \end{subfigure}
    \hfill
    \begin{subfigure}[b]{0.32\linewidth}
        \vspace{0pt}
        \centering
        \includegraphics[width=\linewidth]{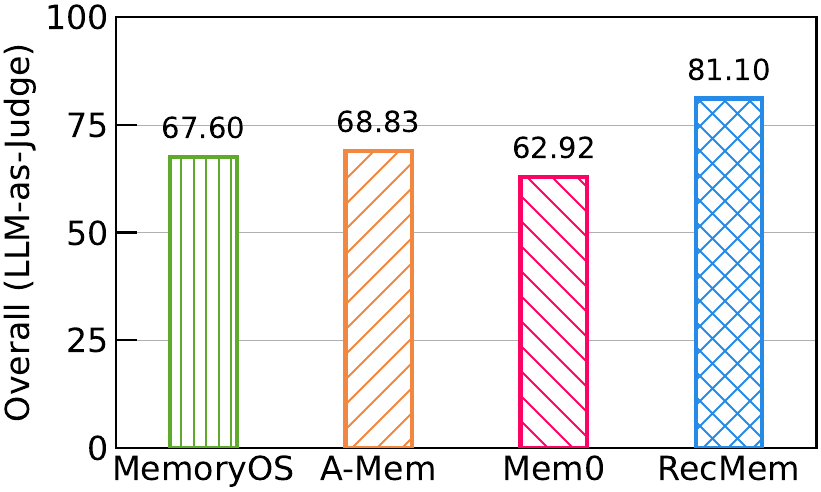}
        \caption{Task accuracy}

        \vspace{0.5em}

        \includegraphics[width=\linewidth]{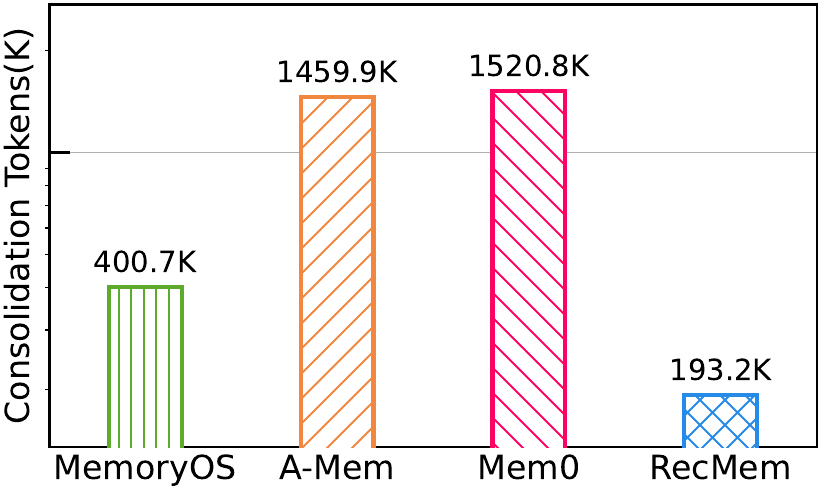}
        \caption{Memory construction cost}
    \end{subfigure}
    \caption{Comparing RecMem with existing memory systems. (a) Existing systems conduct eager memory consolidation for every incoming interaction; (b) our RecMem conducts recurrence-based consolidation selectively from a subconscious memory; (c)-(d) task accuracy and memory construction cost on the LoCoMo benchmark.}
    \label{fig:alpha}
\end{figure*}


Despite the differences in existing memory systems, we observe that they all adopt an \textit{eager memory consolidation} scheme. In particular, for every incoming user-item interaction, they invoke LLMs to extract facts and merge these facts with existing memory contents. This scheme avoids missing information in the interactions but incurs substantial token cost for memory construction, as shown in Figure~\ref{fig:alpha}(d), which makes it expensive to utilize these memory systems in practice. We argue that running LLM-based memory consolidation for every interaction is an overkill. For instance, some interactions may convey little information or contain noise, while some interactions are not related to existing ones and can be queried directly without consolidation. Hence, it is possible to reduce the memory construction cost by choosing when to conduct memory consolidation more judiciously. 


Similar insights emerge from cognitive science. The multi-store theory \citep{memory_model} and the Complementary Learning Systems framework \citep{kumaran,cls,cls2} both converge on a common principle: isolated experiences remain in transient or rapidly-encoded stores, and only repeated or recurring patterns drive consolidation into stable long-term memory. This principle directly motivates RecMem's recurrence-driven consolidation scheme.


Motivated by these insights, we propose RecMem, an efficient memory system for long-running agents that conducts fewer LLM-based memory consolidations in a \textit{recurrence-driven} manner. In particular, RecMem introduces a subconscious memory layer that buffers the raw user-agent interactions via lightweight embeddings, enabling cost-effective retrieval without invoking LLMs. Memory consolidation is only conducted when an incoming interaction can find a sufficient number of semantically similar or related interactions in the subconscious memory, and LLMs are utilized to extract episodic summaries and semantic facts from these interactions. This works because these interactions form a semantic cluster with rich information that is worth memory consolidation and resembles generating long-term memory from transient memory in cognitive science.        




RecMem also incorporates a \emph{semantic refinement} mechanism to improve accuracy. Specifically, LLM-based extraction, especially event-level episodic summarization, may omit fine-grained but query-critical details, leading to lossy long-term memory. Our semantic refinement revisits the raw interactions associated with each episodic memory,  extracts the missing and persistent facts that are not captured by the episodic memory, and distills them into a semantic memory to avoid information loss. 



We empirically evaluate RecMem on two challenging long-term memory benchmarks (i.e., LoCoMo~\cite{locomo} and LongMemEval-S~\cite{longmemeval}) and compare with three SOTA memory systems (i.e., Mem0~\cite{mem0}, A-Mem~\cite{a-mem}, and MemoryOS~\cite{memoryOS}). The results show that RecMem yields higher question answering accuracy than all baselines on both datasets while drastically reducing the token cost for memory construction. In particular, on the LoCoMo benchmark in Figure~\ref{fig:alpha}(d), RecMem reduces the token consumption by up to 7.8x over the baselines. Moreover, RecMem's query-time token cost remains comparable to existing memory systems, so the construction-time savings translate into lower end-to-end cost over long interaction histories.


Our contributions are summarized as follows:
\begin{itemize}[noitemsep,topsep=0pt]
\item We identify a fundamental inefficiency in existing LLM memory systems, i.e., \textit{eager memory consolidation} for every interaction leads to a high memory construction token cost.
\item Inspired by cognitive science, we propose \textit{recurrence-based consolidation} to save token cost by conducting memory consolidation only when an incoming interaction can find a sufficient number of semantically similar or related interactions.
\item We present RecMem, a three-tier memory architecture that realizes this paradigm. Combining a lightweight subconscious store with a novel semantic refinement mechanism, RecMem achieves high accuracy while substantially reducing token cost.
\end{itemize}

\section{Preliminaries}

\subsection{Problem Setting: Conversational Memory}
Recent work on LLM-based agents increasingly focuses on conversational memory, where the agent accumulates information through long-term, multi-turn interactions~\cite{feed_turn,singletomulti,locomo}. Formally, we denote the interaction history available at time step $t$ as a sequence $\mathcal{O}_{1:t}=\{o_1, \ldots, o_t\}$. Each interaction unit $o_t$ is defined as a tuple:
\begin{equation}
    o_t = (s_t, x_t, \tau_t)
\end{equation}
where $s_t \in \{\text{\textsc{user}}, \text{\textsc{assistant}}\}$ represents the speaker role, $x_t$ denotes the message content, and $\tau_t$ is the timestamp. Given a query $q$, the objective is to retrieve relevant evidence from an external memory derived from $\mathcal{O}_{1:t}$ to support reasoning and response generation.

Although conversational settings may appear more specific than general memory scenarios, they capture a fundamental property of real-world deployment: information arrives streamingly over time, and the agent must continually manage an ever-growing interaction history to support future queries and reasoning~\cite{memory_reason}. This formulation contrasts with retrieval-augmented generation (RAG), which typically assumes static or pre-ingested knowledge sources~\cite{Rag,graphrag}. In conversational memory, the key challenge is not retrieval, which can largely leverage existing techniques, but how the system constructs and updates the underlying memory from ongoing interactions in an online manner.

\subsection{Memory Systems}

We focus on training-free, text-based external memory systems for LLM agents in streaming conversational settings. For brevity, we refer to such systems as \emph{memory systems} in the remainder of this paper. Parametric memory approaches~\cite{hippo,m+} require retraining or architectural modification to absorb new information and are thus less applicable in our setting~\cite{feed_turn}, while RL-based methods~\cite{memr1,memalpha} are orthogonal to our focus, as they operate on top of a given memory architecture. 

Most existing memory systems construct long-term memory by incrementally transforming incoming interactions (or short windows thereof) into retrievable memory units, such as summaries~\cite{memoryOS,memgpt,memorybank}, atomic facts~\cite{mem0,mirix}, or structured nodes (e.g., graphs/trees)~\cite{knowledgegraph,zep,treemem}, and then rely on similarity-based retrieval or hybrid search~\cite{memoryOS,zep} to supply evidence at query time. We defer a detailed taxonomy of memory representations, retrieval mechanisms, and construction pipelines to Appendix~\ref{app:related_works}.

\section{The RecMem Framework}
\subsection{Overview}
RecMem is a three-tier memory system guided by the principle that not all interactions warrant LLM-level consolidation. Incoming messages are first organized as atomic interaction units and written to a \emph{subconscious} store with only lightweight structuring and vectorization, making the raw interaction history directly accessible through embedding-based retrieval (\S\ref{sec:subconscious}). Building on this store, RecMem performs \emph{recurrence-based consolidation}: instead of consolidating every turn, it invokes LLM-based processing only when the system observes clear evidence that similar interaction content recurs, thereby reserving LLM invocation for cases where aggregation is likely to be beneficial. Once triggered, RecMem produces an \emph{episodic} abstraction over the selected turns (\S\ref{sec:episodic}), and then applies \emph{semantic refinement} to recover fine-grained, reusable facts that may be omitted by episodic abstraction, grounded in the episode and its underlying interactions (\S\ref{sec:semantic}). At query time, RecMem retrieves a small budget of items from the subconscious, episodic, and semantic stores, and answers by conditioning the LLM on the merged context (\S\ref{sec:answering}). 

Our use of episodic and semantic memory follows the convention in previous LLM memory literature \citep{semantic_episodic_1,mirix}. Specifically, episodic memory in RecMem stores temporally anchored event narratives, which are coherent summaries of how a topic evolves across multiple interaction turns, with explicit time grounding. Semantic memory stores atomic facts about general knowledge, user preferences, constraints, and entity relations.

RecMem's design mirrors human memory: most experiences remain unconsolidated unless repeatedly activated~\cite{memory_model,cls,cls2}. By avoiding eager LLM-based consolidation of transient interactions, RecMem substantially reduces token consumption while preserving both event-level coherence and stable user-centric knowledge as memories. To facilitate understanding, Appendix~\ref{app:example} provides a minimal running example that walks through the memory ingestion workflow.

\subsection{Subconscious Memory}
\label{sec:subconscious}

The subconscious memory manager maintains a faithful record of interaction history at minimal computational cost. A critical design consideration here is the granularity at which conversational information is represented. Existing systems adopt diverse ingestion strategies, ranging from processing individual messages~\cite{a-mem,mirix,zep} or interaction pairs~\cite{mem0} to accumulating larger, fixed-size context buffers~\cite{memoryOS,memgpt}. Static grouping or buffering may conflate temporally adjacent but semantically unrelated topics, diluting the specificity of embeddings. Conversely, ingesting messages in isolation risks fragmenting the semantic context, as an assistant's response often relies heavily on the preceding user query for its meaning.

To address these issues, RecMem treats each \emph{message exchange} (a user-assistant turn) as an atomic unit. Formally, we define a multi-turn conversation between a user and an assistant with $t$ turns as
\begin{align}
\mathcal{H}_t &= \bigl( u_1, u_2, \ldots, u_t \bigr), \\
u_i &= \bigl( m^{\mathrm{usr}}_i, m^{\mathrm{ast}}_i, \tau_i \bigr).
\end{align}

\noindent Here, $u_i$ represents an interaction unit at turn $i$, composed of the user message $m^{\mathrm{usr}}_i$, the assistant response $m^{\mathrm{ast}}_i$, and a timestamp $\tau_i$. The history $\mathcal{H}_t$ is a time-ordered sequence of these units.

As interaction units arrive in a streaming manner, each new message turn $u_i$ is processed independently. We formally define the constructed subconscious memory unit $s_i$ as:
\begin{equation}
s_i = (v_i, u_i) \quad \text{where} \quad v_i = \Phi(u_i).
\end{equation}
These units, computed via the dense vector encoder $\Phi(\cdot)$, are immediately indexed into the subconscious memory store $\mathcal{S}_{\mathrm{sub}}$. This store is implemented as a vector database to support efficient semantic retrieval and incremental updates without the need for batching or access to future context. This fine-grained representation encourages focused semantic embeddings at the level of individual interaction units, making it well-suited for streaming settings.

\paragraph{Recurrence-based Memory Consolidation}
\label{sec:aug}
Echoing cognitive principles~\cite{memory_model,cls,cls2}, we propose \textit{recurrence-based consolidation}: raw interactions are retained in the subconscious buffer, with LLM-based abstraction triggered only when retrieval signals indicate sustained recurrence. Specifically, for a new arriving unit $s_i=(v_i,u_i)$, the system queries $\mathcal{S}_{\mathrm{sub}}$ to retrieve the set $\mathcal{N}_i$ containing the top-$k$ units ranked by cosine similarity to $v_i$. We then filter these candidates to define the \textit{relevant set} based on strict semantic proximity:
\begin{equation}
\mathcal{R}_i = \{\, s_j \in \mathcal{N}_i \mid \cos(v_i, v_j) \ge \theta_{\mathrm{sim}} \}.
\end{equation}

Consolidation is triggered only if the relevant set size meets a recurrence count threshold (i.e., $|\mathcal{R}_i| \ge \theta_{\mathrm{count}}$). In such cases, the cluster $\mathcal{C}_i = \mathcal{R}_i \cup \{s_i\}$ is promoted to higher-level memory modules including episodic memory and semantic memory; otherwise, $s_i$ remains in $\mathcal{S}_{\mathrm{sub}}$. This ensures consolidation is conducted exclusively in memories with demonstrated long-term recurrence.


\subsection{Episodic Memory}
\label{sec:episodic}

Episodic memory captures event-level structure across multiple turns. To ensure memory remains compact, RecMem adopts a merge-first strategy. Upon the arrival of a subconscious unit $s_i=(v_i, u_i)$, we retrieve the nearest neighbor episode $E^\star$ from the episodic store $\mathcal{S}_{\mathrm{epi}}$. Let $v_{E^\star} = \Phi(E^\star)$ denote the embedding of this episode. We strictly enforce an in-place update if semantic similarity permits:
\begin{equation}
\begin{split}
    E^\star &\leftarrow \operatorname{LLM}_{\mathrm{merge}}(E^\star, u_i) \\
    \text{if} \quad &\cos(v_i, v_{E^\star}) \ge \theta_{\mathrm{sim}},
\end{split}
\end{equation}
where $\operatorname{LLM}_{\mathrm{merge}}$ integrates the content of the new turn $u_i$ into the narrative of $E^\star$.

Without such a merge-first step, each recurrence-triggered consolidation on a topic would produce a fresh episode in parallel with existing ones on the same topic, fragmenting the episodic representation of an evolving thread across multiple disconnected entries. Merge-first collapses these into a single continually-updated narrative, keeping the episodic store compact and the per-topic narrative coherent as the conversation evolves.

If merging is not applicable, the unit waits for the \textbf{recurrence-based consolidation} trigger. Given the triggered cluster $\mathcal{C}_i$ (from \S\ref{sec:aug}), we extract the interaction units $\mathcal{U}_{\mathcal{C}} = \{u_j \mid (v_j, u_j) \in \mathcal{C}_i\}$. We then sort these units by their timestamps to form a temporal sequence $U^{\mathrm{seq}}_i = (u^{(1)}, \ldots, u^{(|\mathcal{C}_i|)})$ for episodic memory consolidation. The consolidation prompt is designed for \textit{inductive organization} rather than simple summarization. The LLM processes the formatted sequence to synthesize coherent narratives, segmenting disparate sub-topics if necessary:
\begin{equation}
\mathcal{M}^{\mathrm{epi}}_i = \operatorname{LLM}_{\mathrm{epi}}\left( \bigoplus_{k=1}^{|\mathcal{C}_i|} \operatorname{Fmt}(u^{(k)}) \right).
\end{equation}
Here, $\operatorname{Fmt}(\cdot)$ denotes a fixed template that formats each interaction unit into a textual representation. The output $\mathcal{M}^{\mathrm{epi}}_i$ is a set of new episodic units; each episode $E \in \mathcal{M}^{\mathrm{epi}}_i$ is then encoded via $\Phi(\cdot)$ and stored in $\mathcal{S}_{\mathrm{epi}}$. We provide a complete prompt list in Appendix~\ref{app:prompts} for reference.

\subsection{Semantic Memory}
\label{sec:semantic}

Semantic memory complements episodic memory by storing fine-grained facts that may be missed by event-level summaries. It also mitigates a side effect of the merge-first strategy in episodic memory: as an episode absorbs more turns through repeated merges, its summary necessarily becomes broader and more abstract, which can dilute its retrieval precision for queries that target a specific detail buried within that episode. Semantic memory counteracts this by storing the same details as independent, narrowly-scoped entries, so that precise factual queries can hit them directly without having to surface the entire encompassing episode. In RecMem, we construct this memory layer through a process called \textbf{Semantic Refinement}. By strictly tying semantic extraction to episodic construction, this mechanism ensures that facts remain grounded in the current episodic context while explicitly recovering precise details that were abstracted away.

Formally, when a new episode $E \in \mathcal{M}^{\mathrm{epi}}_i$ is generated from the source interaction units $\mathcal{U}_{\mathcal{C}}$, we first retrieve related existing semantic facts to provide historical context:
\begin{equation}
\mathcal{V} = \operatorname{TopK}_{k}\bigl(\mathcal{S}_{\mathrm{sem}}, \Phi(E)\bigr).
\end{equation}
We then employ an LLM-based refiner to deduce new facts. Conditioned on the raw interaction units $\mathcal{U}_{\mathcal{C}}$, the episodic summary $E$, and the retrieved facts $\mathcal{V}$, the model is instructed to perform two parallel tasks: (1) \textbf{Detail Recovery}, which scans the raw texts in $\mathcal{U}_{\mathcal{C}}$ to identify critical entities omitted by the summary $E$; and (2) \textbf{Fact Maintenance}, which prevents redundancy by filtering out known information in $\mathcal{V}$ while updating evolving user states (e.g., preference changes).

The extraction process is formulated as:
\begin{equation}
\mathcal{M}^{\mathrm{sem}}_i = \operatorname{LLM}_{\mathrm{refine}}\bigl(E, \mathcal{U}_{\mathcal{C}}, \mathcal{V}\bigr).
\end{equation}

Each extracted fact $f \in \mathcal{M}^{\mathrm{sem}}_i$ is stored as an independent entry to preserve retrieval specificity. This design reduces redundancy and enables incremental updates of user facts while keeping retrieval efficient.

\subsection{Question Answering}
\label{sec:answering}
To generate an answer, RecMem first encodes the user query $q$ into a vector representation $v_q = \Phi(q)$ and retrieves the most relevant entries from the subconscious ($\mathcal{S}_{\mathrm{sub}}$), episodic ($\mathcal{S}_{\mathrm{epi}}$), and semantic ($\mathcal{S}_{\mathrm{sem}}$) stores. To manage the context window efficiently while ensuring diverse coverage, we enforce a fixed subconscious retrieval budget and \emph{couple} the episodic and semantic budgets by setting $k_{\mathrm{sem}} = 2k_{\mathrm{epi}}$, yielding three context sets: $\mathcal{K}_{\mathrm{sub}}$, $\mathcal{K}_{\mathrm{epi}}$, and $\mathcal{K}_{\mathrm{sem}}$. The final answer is then generated by conditioning the LLM on the retrieved contexts alongside the original query.

\subsection{Discussions}
\label{subsec:hyperparameter}

\paragraph{Setting the hyper-parameters} RecMem's consolidation behavior is controlled by two key hyper-parameters, i.e., similarity threshold $\theta_{sim}$ for relevant interactions and recurrence threshold $\theta_{count}$ to trigger consolidation. Larger $\theta_{sim}$ and $\theta_{count}$ make consolidation more conservative and favor more frequent patterns, while lower thresholds make consolidation more active and improve coverage for subtle details. According to empirical experiences, we recommend $\theta_{sim}{=}0.7,\ \theta_{count}{=}5$ for casual open-ended settings and $\theta_{sim}{=}0.6,\ \theta_{count}{=}4$ for longer and task-oriented interactions. For question answering, we fix the aggregate retrieval budgets across the memory layers and use $k_{\mathrm{sub}}{=}10$, $k_{\mathrm{epi}}{=}5$, and $k_{\mathrm{sem}}{=}10$ (i.e., $k_{\mathrm{sem}}{=}2k_{\mathrm{epi}}$) by default. Sensitivity experiments for the hyperparameters are conducted in Appendix~\ref{app:hyper}.

\paragraph{Robustness of threshold choice}
A natural concern is whether RecMem's performance depends on precise threshold calibration. Our sensitivity analysis in Appendix~\ref{app:hyper} shows that this is not the case: overall accuracy varies smoothly and is within a narrow band around the recommended defaults, so performance does not hinge on selecting a brittle operating point. Within this robust range, the thresholds instead serve as a strategic dial between memory selectivity and consolidation sensitivity. Higher values of $\theta_{sim}$ and $\theta_{count}$ render RecMem more \textit{conservative}, prioritizing high-confidence patterns suitable for casual open-ended conversations where signal is sparse and noise filtering matters. Conversely, lower thresholds make the system more \textit{active} in consolidation, ideal for task-completion workflows where capturing subtle details is critical.

\paragraph{Generality of recurrence-based consolidation} Although RecMem organizes the episodic memory and semantic memory as flat entries for similarity-based retrieval, recurrence-based consolidation is a general idea and not limited to specific memory structures. The key to recurrence-based consolidation is to utilize a cheap subconscious memory to buffer the incoming interactions and trigger consolidation for higher memory layers based on recurrence, and the higher memory layers can also adopt alternative structures (e.g., knowledge graph).

\section{Experimental Evaluation}

\subsection{Experiment Settings}
To ensure a fair and standardized comparison, we strictly adhere to the incremental evaluation protocol established in prior studies~\cite{mem0,a-mem,memoryOS}. In this setting, message turns are streamed sequentially into the memory system to mimic the natural flow of ongoing dialogues~\cite{feed_turn}, followed by multi-round query sessions. 

\paragraph{Datasets} We evaluate RecMem on two English benchmarks selected to represent distinct interaction modalities: social companionship and long-context task completion. \textbf{LoCoMo}~\cite{locomo} features companion-style, life-sharing dialogues, consisting of 10 multi-session conversations (avg.\ 16k tokens) with questions that probe reasoning over evolving personal history. In contrast, \textbf{LongMemEval-S}~\cite{longmemeval} focuses on agentic, task-oriented interactions with substantially longer contexts. Comprising 500 conversations averaging 115k tokens, it poses a rigorous test for memory systems under realistic, high-load user-assistant workflows. Detailed statistics and question types of these two datasets are provided in Appendix~\ref{app:datasets}.

\paragraph{Baselines}
We compare RecMem against various types of representative baselines: 

\squishlist

    \item \textbf{Full Context}, which feeds all historical interactions to the LLM for answering each question.
    \item  \textbf{Naive RAG}, a standard RAG baseline that segments the interactions into chunks and retrieves the relevant chunks based on embedding similarity. We employ a chunking strategy that respects message integrity (see Appendix~\ref{app:baseline}).
    \item \textbf{Mem0}~\cite{mem0} employs a fact-extraction pipeline to dynamically extract salient information from interactions and manage memory consistency via LLM-based update operations (e.g., add,update, delete).
    \item \textbf{A-Mem}~\cite{a-mem}, an agentic memory system inspired by the Zettelkasten note-taking method~\cite{zettle1,zettle2}, which organizes the interactions as discrete ``memory notes" that are connected via entity linking to facilitate associative retrieval.
    \item \textbf{MemoryOS}~\cite{memoryOS}, an OS-inspired hierarchical framework that manages information via short-term, mid-term, and long-term memory tiers. It also incorporates a dedicated module to maintain evolving user and agent personas to enable personalized interactions.
\squishend

\begin{table*}[!t]
\centering
\small
\setlength{\tabcolsep}{4pt}
\begin{tabular}{p{2.6cm}|cccccc}
\hline
\textbf{Category}
& \textbf{FullContext}
& \textbf{Naive RAG}
& \textbf{MemoryOS}
& \textbf{Mem0}
& \textbf{A-Mem}
& \textbf{RecMem (Ours)} \\
\hline
\multicolumn{6}{l}{\textsc{gpt-4o-mini}}\\
\hline
Multi-Hop        & \underline{69.86} & 54.61 & 58.87 & 52.84 & 52.13 & \textbf{72.70} \\
Temporal         & 52.96 & 24.30 & 44.24 & 52.02 & \textbf{61.99} & \underline{57.32} \\
Open Domain      & \underline{55.21} & 51.04 & 46.88 & 37.50 & 30.21 & \textbf{58.33} \\
Single-Hop       & \textbf{90.01} & 57.55 & 74.55 & 65.52 & 66.83 & \underline{79.79} \\
Overall          & \textbf{76.43} & 49.68 & 63.64 & 58.64 & 60.84 & \underline{72.47} \\
\hline
Construct Token(K) & 0.0 & 0 & 429.7 & 1233.5 & 1143.3 & 202.4 \\
Query Token(K)     & 31.5 & 6.23 & 4.88 & 1.99 & 3.04 & 2.73 \\
\hline
\multicolumn{6}{l}{\textsc{gpt-4.1-mini}}\\
\hline

Multi-Hop        & \textbf{82.62} & 58.87 & 66.31 & 58.16 & 59.93 & \underline{78.37} \\
Temporal         & \textbf{79.13} & 33.96 & 47.66 & 63.86 & 72.90 & \underline{75.39} \\
Open Domain      & \textbf{57.29} & 50.00 & \underline{55.21} & 44.79 & 42.71 & \textbf{57.29} \\
Single-Hop       & \textbf{90.84} & 61.24 & 77.05 & 66.23 & 73.25 & \underline{86.92} \\
Overall          & \textbf{84.18} & 54.42 & 67.60 & 62.92 & 68.83 & \underline{81.10} \\
\hline
Construct Token(K) & 0.00 & 0.00 & 400.7 & 1520.80 & 1459.93 & 193.2 \\
Query Token(K)       & 31.52 & 6.28 & 5.04 & 2.11    & 5.56    & 2.75 \\
\hline
\end{tabular}
\caption{Results on the LoCoMo benchmark. Bold and underline mark the best and second accuracies.}
\label{tab:locomo-main}
\end{table*}


\begin{table*}[!t]
\centering
\small
\setlength{\tabcolsep}{4pt}
\begin{tabular}{p{3.3cm}|cccccc}
\hline
\textbf{Category} & \textbf{FullContext} & \textbf{Naive RAG} & \textbf{MemoryOS} & \textbf{Mem0} & \textbf{A-Mem} & \textbf{RecMem (Ours)} \\
\hline
\multicolumn{6}{l}{\textsc{gpt-4o-mini}}\\
\hline
Single-User & 44.29 & 85.71 & \textbf{97.14} & \underline{95.71} & 92.54 & 91.43 \\
Single-Assistant & 80.36 & 83.93 & \underline{89.29} & 55.36 & \textbf{98.21} & 85.71 \\
Single-Preference & \underline{53.33} & 46.67 & \textbf{70.00} & \textbf{70.00} & 36.67 & 46.67 \\
Temporal-Reasoning & 34.59 & 41.35 & 48.87 & \underline{54.14} & 44.36 & \textbf{61.65} \\
Knowledge-Update & 55.13 & 65.38 & \underline{73.08} & 69.23 & 70.51 & \textbf{80.77} \\
Multi-Session & 35.34 & 52.63 & \textbf{58.65} & \underline{57.89} & 53.38 & 56.39 \\
Overall & 45.60 & 59.40 & \underline{67.8} & 64.00 & 63.80 & \textbf{69.20} \\
\hline
Construct Token(K) & 0.00 & 0.00 & 705.34 & 1244.87 & 1180.23 & 329.55\\
Query Token(K)     & 112.30 & 11.92 & 8.89 & 1.90    & 13.10   & 5.63 \\
\hline
\multicolumn{6}{l}{\textsc{gpt-4.1-mini}}\\
\hline
Single-User & 91.43 & 85.71 & 94.29 & 95.71 & 95.71 & \textbf{95.71} \\
Single-Assistant & \textbf{100.00} & 82.36 & 89.29 & 50.00 & \textbf{100.00} & \underline{96.43} \\
Single-Preference & 56.67 & 83.33 & \textbf{100} & \underline{90.00} & 63.33 & 86.67 \\
Temporal-Reasoning & 48.12 & 45.86 & 54.89 & \underline{62.41} & 52.63 & \textbf{68.42} \\
Knowledge-Update & 75.64 & 75.64 & \underline{80.77} & 74.36 & \textbf{82.05} & 75.64 \\
Multi-Session & 53.38 & 63.91 & \underline{67.67} & \textbf{69.92} & 61.65 & 65.41 \\
Overall & 66.20 & 67.00 & \underline{74.4} & 71.20 & 71.60 & \textbf{76.80} \\
\hline
Construct Token(K) & 0.00 & 0.00 & 669.22 & 1626.54 & 1264.25 & 365.49 \\
Query Token(K)     & 112.25 & 11.93 & 9.19 & 2.00    & 15.46   & 5.89 \\
\hline
\end{tabular}
\caption{Results on the LongMemEval-S benchmark. Bold and underline mark the best and second accuracies.}
\label{tab:longmemeval-main}
\end{table*}

\paragraph{Performance Metrics}
We compare RecMem with the baselines along two dimensions. 

\squishlist

\item \textbf{Question answering accuracy.} We report accuracy as the fraction of questions answered correctly. Following~\cite{mem0}, we use \textit{GPT-4o-mini} as an LLM judge and treat its judgment score as the primary metric. We prioritize this semantic evaluation over token-overlap metrics like F1 score, which can under-estimate correctness for open-ended generation with paraphrases. For completeness, we also report F1 and a comparison in Appendix~\ref{app:F1}. All reported task scores are averaged over three runs. 

\item \textbf{Computation efficiency.} We measure LLM token usage (input plus output) in two phases: (1) \textit{construction cost}, averaged per conversation during memory ingestion, and (2) \textit{query cost}, averaged per question during answering. 

\squishend

\paragraph{Implementation Details}

To evaluate the generalization of our approach, we conduct experiments using two distinct LLM backends: \textit{GPT-4o-mini} and \textit{GPT-4.1-mini}. To ensure fair comparison, for any given benchmark result, RecMem and all baselines share the identical underlying model version. For LLM calls, we set temperature=0.0 and utilize \textit{text-embedding-3-small} for vector embedding generations. For RecMem, we configure the recurrence-based consolidation thresholds to adapt to the distinct interaction densities of each benchmark: $\theta_{{sim}}=0.7, \theta_{{count}}=5$ for LoCoMo, and $\theta_{{sim}}=0.6, \theta_{{count}}=4$ for LongMemEval-S. More detailed experiment settings are  in Appendix~\ref{app:experiment}.

\subsection{Main Results}

Tables~\ref{tab:locomo-main} and~\ref{tab:longmemeval-main} demonstrate that RecMem offers a strong efficiency--performance trade-off compared to prior memory systems. For each task, we highlight the best result in bold and the second-best result with underlining. Across both benchmarks and backbone models, RecMem substantially reduces construction-time token consumption while preserving competitive end-task performance, indicating that eager consolidation is not required to achieve effective long-term memory. We emphasize that our goal is not to dominate every individual task category, but rather to achieve the highest overall accuracy among memory systems under a drastically reduced construction-cost budget.

\paragraph{LoCoMo.}
On LoCoMo with GPT-4.1-mini, RecMem uses only 193.2K construction tokens on average, compared to 1520.8K for Mem0 and 1459.9K for A-Mem, corresponding to reductions of 87.3\% and 86.8\%, respectively. A similar reduction pattern holds for GPT-4o-mini. Despite this drastic decrease in construction cost, RecMem achieves the highest overall score among memory-based methods, indicating that recurrence-based memory consolidation can retain strong long-term memory performance while avoiding the systematic overhead of processing every turn through the LLM. We also note that Full Context slightly outperforms RecMem on LoCoMo, which is consistent with LoCoMo's relatively short conversations (approximately 16K tokens per conversation) where full-context inference remains feasible. However, as shown in table~\ref{tab:longmemeval-main}, this behavior does not generalize to substantially longer settings.

\paragraph{LongMemEval-S.}
A similar but more complex pattern emerges on LongMemEval-S, where conversations are substantially longer and closer to real-world long-lived agents.
With GPT-4.1-mini, RecMem reduces construction tokens by 77.5\% relative to Mem0 and 71.1\% relative to A-Mem, while achieving the best overall score among all evaluated methods, including Full Context and RAG.

At the category level, different systems exhibit complementary strengths, and we do not claim a universal winner across all question types. 
Importantly, RecMem is not designed to dominate every category in isolation; rather, it targets robust \emph{overall} capability with much smaller construction-cost budget. Our results support this goal: despite large reductions in construction tokens, RecMem attains the best overall score on LongMemEval-S.

Beyond the aggregate metric, RecMem's clearest and most consistent gains appear on temporal reasoning, where long-range dependencies are central. We argue this is a structural consequence of recurrence-based consolidation rather than an artifact of tuning. Temporal reasoning requires two capabilities: cross-time linking of co-referent mentions, and reconstructing their chronological order. Eager consolidation systems are disadvantaged on the former: by committing to summary boundaries at each turn or local buffer, they anchor later mentions of an evolving topic to different summaries, fragmenting the thread. RecMem addresses both capabilities by construction: similarity-based clustering in subconscious memory (\S\ref{sec:subconscious}) aggregates co-referent mentions regardless of temporal distance, and timestamp-sorted episodic consolidation (\S\ref{sec:episodic}) reconstructs chronological order within each cluster. Semantic refinement additionally extracts time-anchored facts grounded in the raw interaction units, serving as a second safeguard for fine-grained temporal evidence that episodic abstraction may compress away. 

\paragraph{Construction vs.\ query cost.}
RecMem's efficiency gains comes from reducing construction-time LLM usage. Query-time token consumption stays within a comparable range across memory-based methods under our evaluation protocol because they retrieve a similar order of evidence for answering, whereas construction-time usage diverges sharply depending on how frequently and how heavily a method invokes LLM processing during ingestion. In streaming deployments where new turns arrive continually, these construction-time differences accumulate over time and can dominate total LLM usage, making construction a critical and often overlooked cost driver. 



\subsection{Ablation Study}

\begin{figure}[!t]
    \centering
    \includegraphics[width=\linewidth]{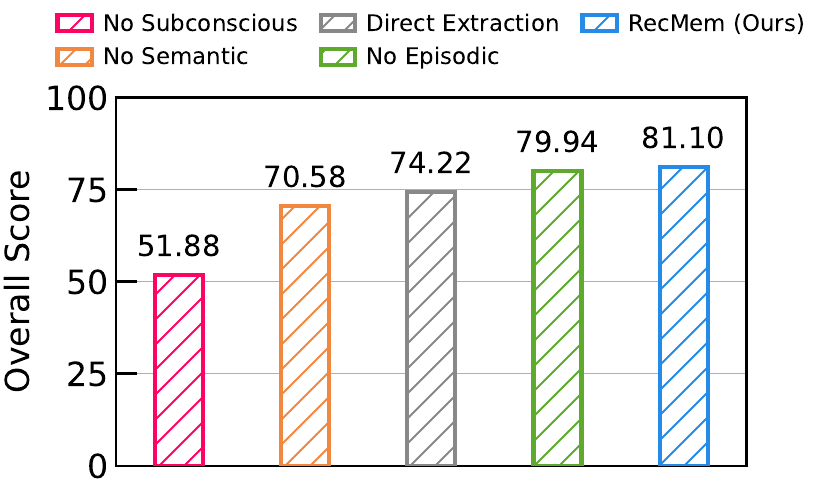}
    \caption{Ablation study for RecMem on LoCoMo}
    \label{fig:locomo-ablation}
\end{figure}

We conduct an ablation study to quantify the contribution of each RecMem module by disabling one component at a time. Figure~\ref{fig:locomo-ablation} reports results on \textbf{LoCoMo} using \textbf{GPT-4.1-mini} as the backbone model. For each testing target, we maintain the retrieval budget for the non-ablated modules to ensure fairness. 

As shown in figure~\ref{fig:locomo-ablation}, overall, removing any module reduces performance, indicating that the three-tier design is complementary. The largest drop occurs when removing subconscious memory ($81.10 \rightarrow 51.88$). This sharp degradation is expected because subconscious memory is the only faithful carrier of raw interaction units: information that does not trigger recurrence-based consolidation remain exclusively in the module, thus disabling it eliminates access to a substantial fraction of query-relevant evidence.

We observe an asymmetric contribution between episodic and semantic memory. Removing episodic memory causes only a small drop ($81.10 \rightarrow 79.94$), whereas removing semantic memory yields a larger but still bounded drop ($81.10 \rightarrow 70.58$). This asymmetry reflects their division of labor under semantic refinement: episodic memories mainly capture high-level structure and cross-turn linkage, while semantic memories prioritize fine-grained factual details. Since semantic refinement explicitly recovers details omitted by episodic abstraction and stores them as semantic facts, semantic memory can partially cover missing evidence when episodic memory is removed; in contrast, episodic summaries can only weakly substitute for the detailed facts lost without semantic memory, leading to the larger degradation.

To isolate the effect of semantic refinement, we evaluate a \textbf{Direct Extraction} variant that extracts semantic facts directly from raw conversations, without using episodic memories as a reference for detecting omitted details. At inference time, this variant answers using only subconscious retrieval and the extracted semantic facts. We remove the refinement-specific guidance tied to episodic summaries in semantic extraction prompt and leave the other parts intact. The score drops from $79.94$ to $74.22$, showing that episodic memory provides an essential reference signal for semantic refinement, improving semantic memory quality beyond naive fact extraction from raw dialogue.

\paragraph{Additional Experiments} Beyond the consolidation thresholds discussed above, we conduct three additional sets of analyses to characterize RecMem's behavior. Appendix~\ref{app:hyper_consolidation} provides a full sensitivity analysis of the consolidation hyperparameters $\theta_{{sim}}$ and $\theta_{{count}}$, examining both accuracy and construction cost. Appendix~\ref{app:hyper_retrieval} studies the retrieval-side budgets $k_{{sub}}$, $k_{{epi}}$, and $k_{{sem}}$ to identify how much evidence is needed at query time. Appendix~\ref{app:F1} additionally reports F1 scores for completeness, along with a discussion of why we treat LLM-as-Judge as the primary metric for open-ended generation.

\section{Conclusion}

We present RecMem, an efficiency-aware memory system for long-running LLM agents that challenges the prevailing paradigm of eager memory consolidation. By explicitly modeling raw interactions within a lightweight \textit{subconscious memory} and deferring LLM-based abstraction until triggered by recurrence, RecMem demonstrates that high-fidelity long-term memory does not necessitate exhaustive processing of every interaction. Across LoCoMo and LongMemEval-S, this strategy substantially reduces memory construction cost while preserving competitive task performance. More broadly, RecMem reframes memory consolidation as a dynamic, recurrence-driven process. We hope this work encourages the community to reconsider \textit{when} and \textit{why} information should be consolidated in long-running agent tasks, and to treat computational cost as a first-class criterion when evaluating future memory systems.

\section{Limitations}

Despite the empirical strengths and efficiency gains of RecMem, several limitations merit discussion.

\paragraph{Dependence on Heuristic Thresholds.}
RecMem relies on static similarity ($\theta_{sim}$) and recurrence thresholds ($\theta_{count}$) to govern the consolidation process. While our experiments demonstrate that these parameters can be tuned to accommodate different interaction densities (e.g., casual conversation vs. task completion), they currently remain manually specified. This dependency means RecMem may benefit from threshold recalibration when deploying to domains with substantially different interaction densities, although Appendix $\ref{app:hyper}$ shows that such recalibration can be coarse rather than precise. Developing adaptive or learnable triggering mechanisms that dynamically adjust to user behavior is a promising direction for future work.

\paragraph{Recurrence as a Proxy for Salience.}
Our design is predicated on the assumption that information worthy of long-term abstraction tends to recur. While this aligns with many cognitive theories and conversational patterns, it may risk overlooking rare but critical events—such as a one-off safety instruction or a unique user constraint—that appear only once. To mitigate this risk, the subconscious memory layer functions as a persistent safety net: every interaction unit is preserved verbatim and remains directly retrievable at query time, regardless of whether it has been consolidated. Nevertheless, non-recurring content does not benefit from the cross-turn linking of episodic memory or the fact-level refinement of semantic memory, which may weaken reasoning over these details. Developing a lightweight salience signal beyond pure recurrence to promote rare but high-value events is a promising direction for future work.

\section{Ethical Considerations}

We evaluate RecMem only on publicly available benchmarks in an offline setting, and we do not deploy or test it in real user-facing applications. Nevertheless, long-term memory mechanisms can raise dual-use concerns: when integrated into real applications, persistent memory may be misused for profiling or surveillance beyond the intended personalization benefits. We therefore recommend that practical deployments incorporate clear user-facing disclosures and safeguards such as access controls and user-controllable deletion/retention policies.

A second risk arises from unintended harms due to incorrect memory. Errors in consolidation or retrieval can surface outdated or spurious details and lead to overconfident but incorrect responses, which may be consequential in high-stakes settings. We encourage future work to incorporate uncertainty-aware retrieval, confidence calibration, and monitoring against memory poisoning or prompt-injection attempts.

Finally, RecMem reduces unnecessary LLM invocations compared to eager extraction baselines, which can lower compute and associated environmental footprint when operating over long interaction histories.
\bibliography{custom}

\begin{thebibliography}{33}
\providecommand{\natexlab}[1]{#1}

\bibitem[{Ahrens(2017)}]{zettle2}
S.~Ahrens. 2017.
\newblock \href {https://books.google.com.sg/books?id=lHDsDwAAQBAJ} {\emph{How to Take Smart Notes: One Simple Technique to Boost Writing, Learning and Thinking -- for Students, Academics and Nonfiction Book Writers}}.
\newblock S{\"o}nke Ahrens.

\bibitem[{Atkinson and Shiffrin(1968)}]{memory_model}
Richard~C. Atkinson and Richard~M. Shiffrin. 1968.
\newblock \href {https://api.semanticscholar.org/CorpusID:22958289} {Human memory: A proposed system and its control processes}.
\newblock In \emph{The psychology of learning and motivation}.

\bibitem[{Chhikara et~al.(2025)Chhikara, Khant, Aryan, Singh, and Yadav}]{mem0}
Prateek Chhikara, Dev Khant, Saket Aryan, Taranjeet Singh, and Deshraj Yadav. 2025.
\newblock \href {https://arxiv.org/abs/2504.19413} {Mem0: Building production-ready ai agents with scalable long-term memory}.
\newblock \emph{Preprint}, arXiv:2504.19413.

\bibitem[{Fang et~al.(2025)Fang, Yu, Zhong, Ye, Xiong, and Wei}]{hippo}
Yunhao Fang, Weihao Yu, Shu Zhong, Qinghao Ye, Xuehan Xiong, and Lai Wei. 2025.
\newblock \href {https://arxiv.org/abs/2510.07318} {Artificial hippocampus networks for efficient long-context modeling}.
\newblock \emph{Preprint}, arXiv:2510.07318.

\bibitem[{Guo et~al.(2024)Guo, Zhu, Yang, Xie, Dong, Zhang, Chen, Bi, Wu, Li, Luo, Xiong, and Liang}]{deepseek_code}
Daya Guo, Qihao Zhu, Dejian Yang, Zhenda Xie, Kai Dong, Wentao Zhang, Guanting Chen, Xiao Bi, Y.~Wu, Y.~K. Li, Fuli Luo, Yingfei Xiong, and Wenfeng Liang. 2024.
\newblock \href {https://arxiv.org/abs/2401.14196} {Deepseek-coder: When the large language model meets programming -- the rise of code intelligence}.
\newblock \emph{Preprint}, arXiv:2401.14196.

\bibitem[{Han et~al.(2025)Han, Wang, Shomer, Guo, Ding, Lei, Halappanavar, Rossi, Mukherjee, Tang, He, Hua, Long, Zhao, Shah, Javari, Xia, and Tang}]{graphrag}
Haoyu Han, Yu~Wang, Harry Shomer, Kai Guo, Jiayuan Ding, Yongjia Lei, Mahantesh Halappanavar, Ryan~A. Rossi, Subhabrata Mukherjee, Xianfeng Tang, Qi~He, Zhigang Hua, Bo~Long, Tong Zhao, Neil Shah, Amin Javari, Yinglong Xia, and Jiliang Tang. 2025.
\newblock \href {https://arxiv.org/abs/2501.00309} {Retrieval-augmented generation with graphs (graphrag)}.
\newblock \emph{Preprint}, arXiv:2501.00309.

\bibitem[{Hogan et~al.(2021)Hogan, Blomqvist, Cochez, D’amato, Melo, Gutierrez, Kirrane, Gayo, Navigli, Neumaier, Ngomo, Polleres, Rashid, Rula, Schmelzeisen, Sequeda, Staab, and Zimmermann}]{knowledgegraph}
Aidan Hogan, Eva Blomqvist, Michael Cochez, Claudia D’amato, Gerard~De Melo, Claudio Gutierrez, Sabrina Kirrane, José Emilio~Labra Gayo, Roberto Navigli, Sebastian Neumaier, Axel-Cyrille~Ngonga Ngomo, Axel Polleres, Sabbir~M. Rashid, Anisa Rula, Lukas Schmelzeisen, Juan Sequeda, Steffen Staab, and Antoine Zimmermann. 2021.
\newblock \href {https://doi.org/10.1145/3447772} {Knowledge graphs}.
\newblock \emph{ACM Computing Surveys}, 54(4):1–37.

\bibitem[{Hu et~al.(2025)Hu, Wang, and McAuley}]{feed_turn}
Yuanzhe Hu, Yu~Wang, and Julian McAuley. 2025.
\newblock \href {https://arxiv.org/abs/2507.05257} {Evaluating memory in llm agents via incremental multi-turn interactions}.
\newblock \emph{Preprint}, arXiv:2507.05257.

\bibitem[{Jiang et~al.(2025)Jiang, Li, Zhao, Qiu, Wang, Shao, Xu, Zhang, Chen, Tang, Chen, Wu, Ma, Wang, and Chen}]{self_evolve}
Xun Jiang, Feng Li, Han Zhao, Jiahao Qiu, Jiaying Wang, Jun Shao, Shihao Xu, Shu Zhang, Weiling Chen, Xavier Tang, Yize Chen, Mengyue Wu, Weizhi Ma, Mengdi Wang, and Tianqiao Chen. 2025.
\newblock \href {https://arxiv.org/abs/2410.15665} {Long term memory: The foundation of ai self-evolution}.
\newblock \emph{Preprint}, arXiv:2410.15665.

\bibitem[{Johnson et~al.(2017)Johnson, Douze, and Jégou}]{faiss}
Jeff Johnson, Matthijs Douze, and Hervé Jégou. 2017.
\newblock \href {https://arxiv.org/abs/1702.08734} {Billion-scale similarity search with gpus}.
\newblock \emph{Preprint}, arXiv:1702.08734.

\bibitem[{Kadavy(2021)}]{zettle1}
David Kadavy. 2021.
\newblock \emph{Digital Zettelkasten: Principles, Methods, \& Examples}.
\newblock Kadavy, Incorporated.

\bibitem[{Kang et~al.(2025)Kang, Ji, Zhao, and Bai}]{memoryOS}
Jiazheng Kang, Mingming Ji, Zhe Zhao, and Ting Bai. 2025.
\newblock \href {https://arxiv.org/abs/2506.06326} {Memory os of ai agent}.
\newblock \emph{Preprint}, arXiv:2506.06326.

\bibitem[{Kumaran et~al.(2016)Kumaran, Hassabis, and McClelland}]{kumaran}
Dharshan Kumaran, Demis Hassabis, and James~L. McClelland. 2016.
\newblock \href {https://doi.org/10.1016/j.tics.2016.05.004} {What learning systems do intelligent agents need? complementary learning systems theory updated}.
\newblock \emph{Trends in Cognitive Sciences}, 20(7):512--534.

\bibitem[{Lewis et~al.(2021)Lewis, Perez, Piktus, Petroni, Karpukhin, Goyal, Küttler, Lewis, tau Yih, Rocktäschel, Riedel, and Kiela}]{Rag}
Patrick Lewis, Ethan Perez, Aleksandra Piktus, Fabio Petroni, Vladimir Karpukhin, Naman Goyal, Heinrich Küttler, Mike Lewis, Wen tau Yih, Tim Rocktäschel, Sebastian Riedel, and Douwe Kiela. 2021.
\newblock \href {https://arxiv.org/abs/2005.11401} {Retrieval-augmented generation for knowledge-intensive nlp tasks}.
\newblock \emph{Preprint}, arXiv:2005.11401.

\bibitem[{Li and Li(2024)}]{semantic_episodic_1}
Jitang Li and Jinzheng Li. 2024.
\newblock \href {https://arxiv.org/abs/2401.02509} {Memory, consciousness and large language model}.
\newblock \emph{Preprint}, arXiv:2401.02509.

\bibitem[{Liu et~al.(2025)Liu, Li, Zhang, Wang, He, Hong, Liu, Zhang, Song, Zhu, Cheng, Wang, Wang, Luo, Jin, Zhang, Liu, Chen, Zhang, Yu, Shi, Li, Wu, Teng, Jia, Xu, Xiang, Lin, Liu, Liu, Su, Sun, Berseth, Nie, Foster, Ward, Wu, Gu, Zhuge, Liang, Tang, Wang, You, Wang, Pei, Yang, Qi, and Wu}]{forget}
Bang Liu, Xinfeng Li, Jiayi Zhang, Jinlin Wang, Tanjin He, Sirui Hong, Hongzhang Liu, Shaokun Zhang, Kaitao Song, Kunlun Zhu, Yuheng Cheng, Suyuchen Wang, Xiaoqiang Wang, Yuyu Luo, Haibo Jin, Peiyan Zhang, Ollie Liu, Jiaqi Chen, Huan Zhang, and 29 others. 2025.
\newblock \href {https://arxiv.org/abs/2504.01990} {Advances and challenges in foundation agents: From brain-inspired intelligence to evolutionary, collaborative, and safe systems}.
\newblock \emph{Preprint}, arXiv:2504.01990.

\bibitem[{Liu et~al.(2023)Liu, Lin, Hewitt, Paranjape, Bevilacqua, Petroni, and Liang}]{lostinthemiddle}
Nelson~F. Liu, Kevin Lin, John Hewitt, Ashwin Paranjape, Michele Bevilacqua, Fabio Petroni, and Percy Liang. 2023.
\newblock \href {https://api.semanticscholar.org/CorpusID:259360665} {Lost in the middle: How language models use long contexts}.
\newblock \emph{Transactions of the Association for Computational Linguistics}, 12:157--173.

\bibitem[{Maharana et~al.(2024)Maharana, Lee, Tulyakov, Bansal, Barbieri, and Fang}]{locomo}
Adyasha Maharana, Dong-Ho Lee, Sergey Tulyakov, Mohit Bansal, Francesco Barbieri, and Yuwei Fang. 2024.
\newblock \href {https://arxiv.org/abs/2402.17753} {Evaluating very long-term conversational memory of llm agents}.
\newblock \emph{Preprint}, arXiv:2402.17753.

\bibitem[{McClelland et~al.(1995)McClelland, McNaughton, and O’Reilly}]{cls2}
James~L. McClelland, Bruce~L. McNaughton, and Randall~C. O’Reilly. 1995.
\newblock \href {https://api.semanticscholar.org/CorpusID:2832081} {Why there are complementary learning systems in the hippocampus and neocortex: insights from the successes and failures of connectionist models of learning and memory.}
\newblock \emph{Psychological review}, 102 3:419--457.

\bibitem[{O’Reilly et~al.(2014)O’Reilly, Bhattacharyya, Howard, and Ketz}]{cls}
Randall~C. O’Reilly, Rajan Bhattacharyya, Michael~D. Howard, and Nicholas Ketz. 2014.
\newblock \href {https://doi.org/10.1111/j.1551-6709.2011.01214.x} {Complementary learning systems}.
\newblock \emph{Cognitive Science}, 38(6):1229--1248.

\bibitem[{Packer et~al.(2024)Packer, Wooders, Lin, Fang, Patil, Stoica, and Gonzalez}]{memgpt}
Charles Packer, Sarah Wooders, Kevin Lin, Vivian Fang, Shishir~G. Patil, Ion Stoica, and Joseph~E. Gonzalez. 2024.
\newblock \href {https://arxiv.org/abs/2310.08560} {Memgpt: Towards llms as operating systems}.
\newblock \emph{Preprint}, arXiv:2310.08560.

\bibitem[{Rasmussen et~al.(2025)Rasmussen, Paliychuk, Beauvais, Ryan, and Chalef}]{zep}
Preston Rasmussen, Pavlo Paliychuk, Travis Beauvais, Jack Ryan, and Daniel Chalef. 2025.
\newblock \href {https://arxiv.org/abs/2501.13956} {Zep: A temporal knowledge graph architecture for agent memory}.
\newblock \emph{Preprint}, arXiv:2501.13956.

\bibitem[{Rezazadeh et~al.(2025)Rezazadeh, Li, Wei, and Bao}]{treemem}
Alireza Rezazadeh, Zichao Li, Wei Wei, and Yujia Bao. 2025.
\newblock \href {https://arxiv.org/abs/2410.14052} {From isolated conversations to hierarchical schemas: Dynamic tree memory representation for llms}.
\newblock \emph{Preprint}, arXiv:2410.14052.

\bibitem[{Shao et~al.(2024)Shao, Wang, Zhu, Xu, Song, Bi, Zhang, Zhang, Li, Wu, and Guo}]{deepseek_math}
Zhihong Shao, Peiyi Wang, Qihao Zhu, Runxin Xu, Junxiao Song, Xiao Bi, Haowei Zhang, Mingchuan Zhang, Y.~K. Li, Y.~Wu, and Daya Guo. 2024.
\newblock \href {https://arxiv.org/abs/2402.03300} {Deepseekmath: Pushing the limits of mathematical reasoning in open language models}.
\newblock \emph{Preprint}, arXiv:2402.03300.

\bibitem[{Wang and Chen(2025)}]{mirix}
Yu~Wang and Xi~Chen. 2025.
\newblock \href {https://arxiv.org/abs/2507.07957} {Mirix: Multi-agent memory system for llm-based agents}.
\newblock \emph{Preprint}, arXiv:2507.07957.

\bibitem[{Wang et~al.(2025{\natexlab{a}})Wang, Krotov, Hu, Gao, Zhou, McAuley, Gutfreund, Feris, and He}]{m+}
Yu~Wang, Dmitry Krotov, Yuanzhe Hu, Yifan Gao, Wangchunshu Zhou, Julian McAuley, Dan Gutfreund, Rogerio Feris, and Zexue He. 2025{\natexlab{a}}.
\newblock \href {https://arxiv.org/abs/2502.00592} {M+: Extending memoryllm with scalable long-term memory}.
\newblock \emph{Preprint}, arXiv:2502.00592.

\bibitem[{Wang et~al.(2025{\natexlab{b}})Wang, Takanobu, Liang, Mao, Hu, McAuley, and Wu}]{memalpha}
Yu~Wang, Ryuichi Takanobu, Zhiqi Liang, Yuzhen Mao, Yuanzhe Hu, Julian McAuley, and Xiaojian Wu. 2025{\natexlab{b}}.
\newblock \href {https://arxiv.org/abs/2509.25911} {Mem-$\alpha$: Learning memory construction via reinforcement learning}.
\newblock \emph{Preprint}, arXiv:2509.25911.

\bibitem[{Wu et~al.(2025)Wu, Wang, Yu, Zhang, Chang, and Yu}]{longmemeval}
Di~Wu, Hongwei Wang, Wenhao Yu, Yuwei Zhang, Kai-Wei Chang, and Dong Yu. 2025.
\newblock \href {https://arxiv.org/abs/2410.10813} {Longmemeval: Benchmarking chat assistants on long-term interactive memory}.
\newblock \emph{Preprint}, arXiv:2410.10813.

\bibitem[{Xu et~al.(2025{\natexlab{a}})Xu, Wen, Jia, Zhang, wenlin zhang, Wang, Guo, Tang, Zhao, Chen, and Xu}]{singletomulti}
Derong Xu, Yi~Wen, Pengyue Jia, Yingyi Zhang, wenlin zhang, Yichao Wang, Huifeng Guo, Ruiming Tang, Xiangyu Zhao, Enhong Chen, and Tong Xu. 2025{\natexlab{a}}.
\newblock \href {https://arxiv.org/abs/2505.19549} {From single to multi-granularity: Toward long-term memory association and selection of conversational agents}.
\newblock \emph{Preprint}, arXiv:2505.19549.

\bibitem[{Xu et~al.(2025{\natexlab{b}})Xu, Liang, Mei, Gao, Tan, and Zhang}]{a-mem}
Wujiang Xu, Zujie Liang, Kai Mei, Hang Gao, Juntao Tan, and Yongfeng Zhang. 2025{\natexlab{b}}.
\newblock \href {https://arxiv.org/abs/2502.12110} {A-mem: Agentic memory for llm agents}.
\newblock \emph{Preprint}, arXiv:2502.12110.

\bibitem[{Yan et~al.(2025)Yan, Yang, Huang, Nie, Ding, Li, Ma, Kersting, Pan, Schütze, Tresp, and Ma}]{memr1}
Sikuan Yan, Xiufeng Yang, Zuchao Huang, Ercong Nie, Zifeng Ding, Zonggen Li, Xiaowen Ma, Kristian Kersting, Jeff~Z. Pan, Hinrich Schütze, Volker Tresp, and Yunpu Ma. 2025.
\newblock \href {https://arxiv.org/abs/2508.19828} {Memory-r1: Enhancing large language model agents to manage and utilize memories via reinforcement learning}.
\newblock \emph{Preprint}, arXiv:2508.19828.

\bibitem[{Zhang et~al.(2024)Zhang, Bo, Ma, Li, Chen, Dai, Zhu, Dong, and Wen}]{memory_reason}
Zeyu Zhang, Xiaohe Bo, Chen Ma, Rui Li, Xu~Chen, Quanyu Dai, Jieming Zhu, Zhenhua Dong, and Ji-Rong Wen. 2024.
\newblock \href {https://arxiv.org/abs/2404.13501} {A survey on the memory mechanism of large language model based agents}.
\newblock \emph{Preprint}, arXiv:2404.13501.

\bibitem[{Zhong et~al.(2023)Zhong, Guo, Gao, Ye, and Wang}]{memorybank}
Wanjun Zhong, Lianghong Guo, Qiqi Gao, He~Ye, and Yanlin Wang. 2023.
\newblock \href {https://arxiv.org/abs/2305.10250} {Memorybank: Enhancing large language models with long-term memory}.
\newblock \emph{Preprint}, arXiv:2305.10250.

\end{thebibliography}
\appendix
\section{Detailed Taxonomy of Memory Systems}
\label{app:related_works}

In this section, we provide a structured taxonomy of existing memory systems, focusing on two critical dimensions: memory consolidation (how raw interactions are transformed into long-term storage) and retrieval mechanisms (how relevant information is accessed during query time).

\subsection{Memory Consolidation Paradigms}

Memory consolidation transforms raw interaction streams into retrievable long-term storage, as described in \S\ref{sec:introduction}. A critical commonality across existing works is their reliance on an eager consolidation strategy. In these systems, every incoming interaction—regardless of its informational value or redundancy—eventually triggers an LLM-driven processing pipeline. This approach assumes that all user inputs require active structuring or abstraction, incurring constant computational overhead to maintain the memory state. We categorize these paradigms by their consolidation targets:

\paragraph{Graph and Structure-based Consolidation.}
These systems treat memory construction as a continuous structural maintenance task. Upon receiving a new message, the system must compute embeddings, identify entities, and execute structural updates (e.g., creating nodes or re-balancing trees) to integrate the new information into the existing topology.

\begin{enumerate}
    \item A-Mem~\cite{a-mem}: Inspired by the Zettelkasten method~\cite{zettle1,zettle2}, it treats interactions as discrete "notes" in a network, where consolidation involves generating embeddings and establishing associative links between new and existing notes.
    \item TreeMem~\cite{treemem}: Maintains a hierarchical summary tree. New information is not just appended but traverses down to specific leaf nodes based on semantic relevance, forcing a recursive chain of summary updates from the leaf back up to the root to keep the hierarchy consistent.
    \item Zep~\cite{zep}: Parses interactions into a "Temporal Knowledge Graph." It actively extracts entities and relationships from each turn, modeling them as nodes and edges while explicitly updating the temporal metadata of these connections.
    \item Mem0 (Graph Variant)~\cite{mem0}: Extends atomic fact extraction by organizing data into a graph. It requires per-turn analysis to identify multi-hop relationships between entities, dynamically updating the graph structure as the conversation evolves.
\end{enumerate}

\paragraph{Fact and Summary-based Consolidation}
These systems function as active distillers, where the LLM is invoked at every turn (or small buffer intervals) to parse information into compressed formats. The goal is to immediately strip away redundancy and store only the event summaries or extracted facts.

\begin{enumerate}
    \item Mem0~\cite{mem0}: Runs a dedicated extraction pipeline after every user message. It prompts the LLM to identify atomic facts (e.g., entity-relation triplets), instructing it to add, update, or delete records in the vector database to reflect the latest state.
    \item MemoryOS~\cite{memoryOS}: Features a multi-tiered architecture (Short-, Mid-, and Long-term memories) to manage context flow, emphasizing a dedicated Profile Memory module that explicitly maintains evolving user personas and agent guidelines.
    \item Mirix~\cite{mirix}: Routes every interaction through a parallel extraction pipeline. Raw text is simultaneously processed by distinct modules to distill specific "Knowledge" facts and "Event" summaries, creating a synchronized update across multiple memory stores.
    \item MemGPT~\cite{memgpt}: Treats memory management as an operating system process, employing self-directed function calls to actively summarize and compress ongoing interactions into a fixed-size "Core Memory" block, ensuring key persona and user details are preserved while offloading raw history. 
\end{enumerate}

\subsection{Retrieval Mechanisms}

While memory consolidation determines how information is stored, retrieval mechanisms define how relevant context is accessed to support reasoning. Existing approaches range from simple semantic matching to complex, structure-aware traversal algorithms.

\paragraph{Dense Vector Retrieval}
This prevalent paradigm relies on high-dimensional embeddings to measure semantic overlap, commonly utilizing vector databases like FAISS~\cite{faiss} for efficient similarity search. A representative system is Mem0~\cite{mem0}, which retrieves relevant atomic facts by computing the cosine similarity between the query and stored embeddings, selecting the top-$k$ entries based purely on semantic relevance scores.
\begin{figure*}[ht]
    \centering
    \includegraphics[width=\linewidth]{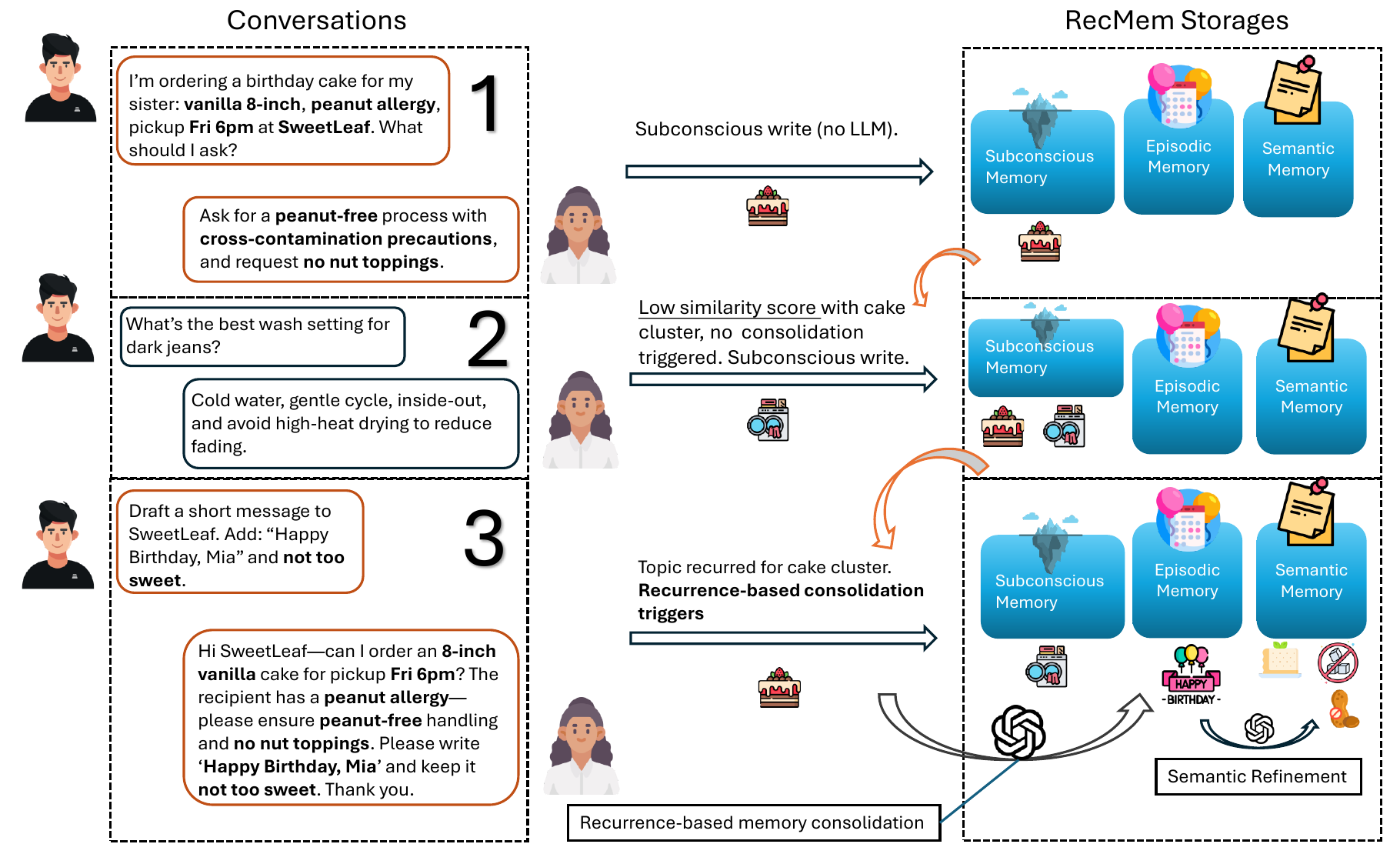}
    \caption{A simplified memory ingestion process in RecMem}
    \label{fig:Example}
\end{figure*}
\paragraph{Structure-Aware Retrieval} These systems leverage the topological structure established during consolidation (graphs or trees) to expand retrieval beyond simple similarity. TreeMem~\cite{treemem} utilizes a top-down tree pruning strategy; starting from the root, it evaluates child nodes based on their summaries and prunes irrelevant branches to efficiently narrow the search to specific leaf nodes. Similarly, A-Mem~\cite{a-mem} employs associative retrieval: upon locating an initial "note" via vector search, it traverses established entity links to fetch connected notes, mimicking the human ability to associate disparate memories through shared concepts.

\paragraph{Hybrid Retrieval} To mitigate the precision limitations of pure vector search (e.g., missing exact keyword matches), some systems adopt a multi-metric strategy. MemoryOS~\cite{memoryOS} implements a weighted hybrid retrieval mechanism. Instead of relying on a single metric, it calculates a unified relevance score by linearly combining Cosine similarity (for semantic understanding) and Jaccard similarity (for exact keyword overlap). This approach ensures that specific entities are recalled even if their semantic embeddings are distant, balancing fuzzy semantic matching with precise lexical matching.

\section{Running Example}
\label{app:example}
We briefly illustrate RecMem's ingestion-time behavior with a minimal three-turn interaction in Figure~\ref{fig:Example}. For clarity, we set the recurrence threshold to $\theta_{\text{count}}=2$: consolidation is triggered once a topic is observed in at least two interaction units after passing the topical-similarity check.
For simplicity, we do not expand the exact similarity threshold here and use natural language describe when two turns are treated as relevant.
To keep the example concise, we present only the recurrence-triggered construction path, and therefore omit the \textbf{merge-first} episodic in-place update.

\paragraph{Turn 1: subconscious write without consolidation.}
The user first asks for suggestions to order a birthday cake.
RecMem ingests this user--assistant exchange as one interaction unit and appends it to the subconscious memory.
Since the ``cake'' topic has been observed only once so far, the corresponding set $R_i$ does not satisfy the recurrence condition $|R_i|\ge \theta_{\text{count}}$, and thus no LLM-based consolidation is triggered. Instead, RecMem computes a lightweight embedding for this unit and stores it together with the raw text in the subconscious vector index, enabling efficient similarity-based retrieval in future turns.
\begin{figure*}[ht]
    \centering
    \begin{subfigure}[t]{0.31\linewidth}
        \vspace{0pt} 
        \includegraphics[width=\linewidth]{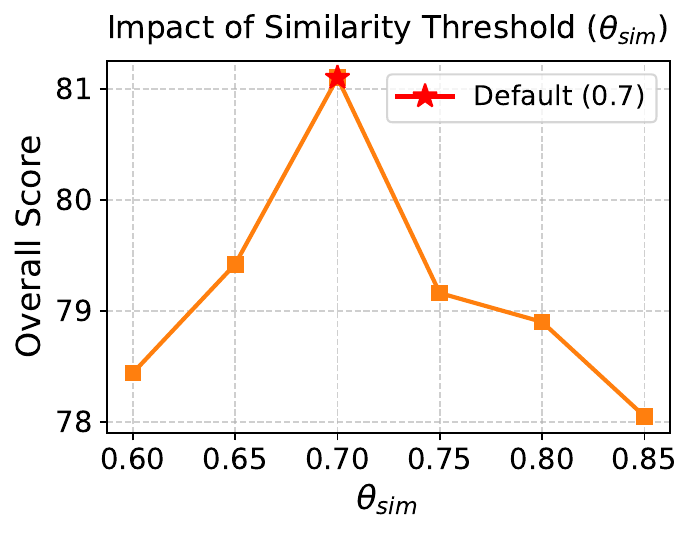}
        \caption{}
        \label{hyper:sim}
    \end{subfigure}
    \hfill
    \begin{subfigure}[t]{0.32\linewidth}
        \vspace{0pt} 
        \includegraphics[width=\linewidth]{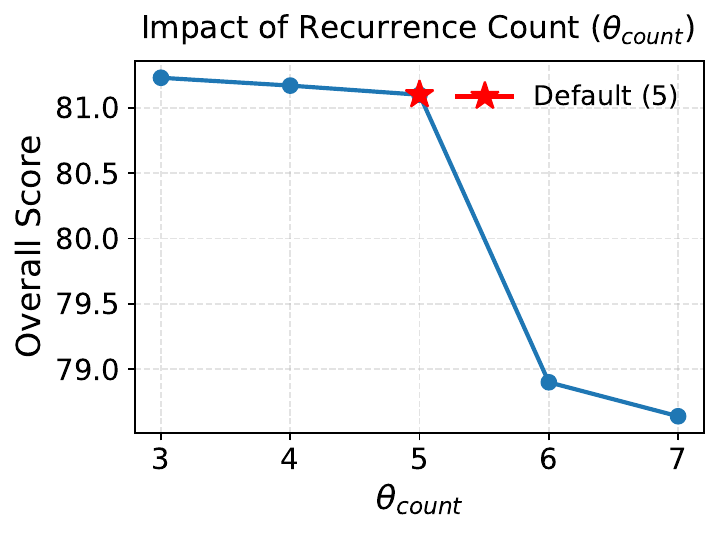}
        \caption{}
        \label{hyper:count}
    \end{subfigure}
    \hfill
    \begin{subfigure}[t]{0.32\linewidth}
        \vspace{0pt}
        \centering
        \includegraphics[width=\linewidth]{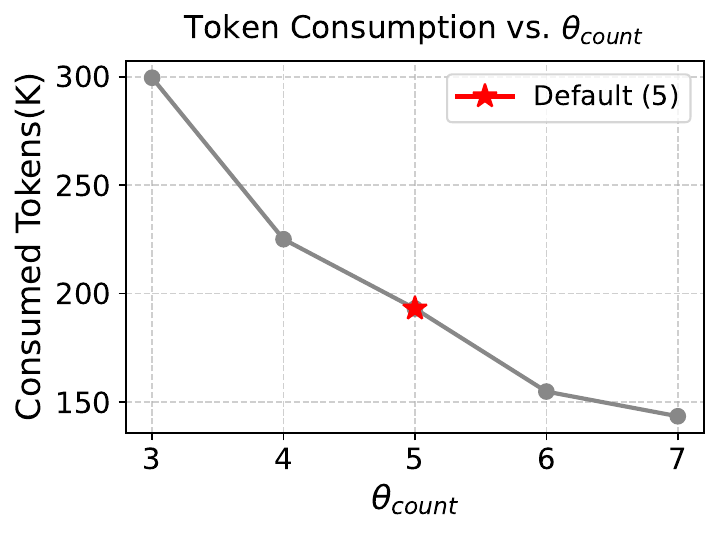}
        \caption{}
        \label{hyper:count_cost}
    \end{subfigure}
    \caption{Sensitivity of consolidation thresholds on LoCoMo (GPT-4.1-mini). (a) Overall score vs. $\theta_{\mathrm{sim}}$. (b) Overall score vs. $\theta_{\mathrm{count}}$. (c) Memory-construction token consumption vs. $\theta_{\mathrm{count}}$.}
    \label{fig:hyper_consolidation}
\end{figure*}
\paragraph{Turn 2: no recurrence under the similarity check.}
The user then switches to an unrelated topic (washing dark jeans).
RecMem uses the new unit's embedding to retrieve relevant turns from the subconscious store, and then forms $R_i$ by keeping only those with similarity above the topical threshold.
The cake-related unit from Turn~1 is unrelated to this turn thus $|R_i|=1$ and consolidation is not triggered. The new turn is then stored in subconscious store.

\paragraph{Turn 3: recurrence-based consolidation and semantic refinement.}
When the user returns to the cake topic, the new unit retrieves prior cake-related unit(s) and passes the similarity check $\theta_{\text{sim}}$.
Since the recurrence count now satisfies $|R_i|\ge \theta_{\text{count}}$, RecMem triggers consolidation and produces two complementary artifacts.
\textbf{Episodic memory} abstracts the recurring turns into a coherent, intent-level narrative---e.g., the user is preparing a birthday cake order for their sister Mia, and the assistant recommends an allergy-safe ordering strategy. This abstraction focuses on high-level event summary but may compress away valuable details.
\textbf{Semantic refinement} preserves such details by extracting atomic facts from the underlying raw turns, such as: (i) the user has a sister named Mia; (ii) Mia is allergic to peanuts; and (iii) the user plans to place an order at SweetLeaf with concrete cake/message specifications. Semantic refinement also uses related existing semantic memories to assist extraction, but we omit this aspect here to keep the example minimal.

\section{Hyperparameter Analysis}
\label{app:hyper}
In this section, we conduct a sensitivity analysis of RecMem's key hyperparameters under a controlled-variable protocol.
\begin{figure*}[ht]
    \centering
    \begin{subfigure}[t]{0.45\linewidth}
        \vspace{0pt} 
        \includegraphics[width=\linewidth]{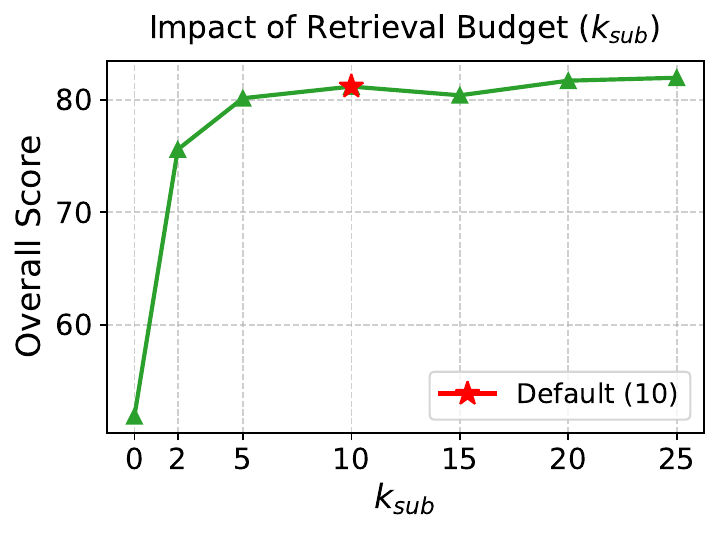}
        \caption{}
        \label{hyper:sub}
    \end{subfigure}
    \hfill
    \begin{subfigure}[t]{0.45\linewidth}
        \vspace{0pt} 
        \includegraphics[width=\linewidth]{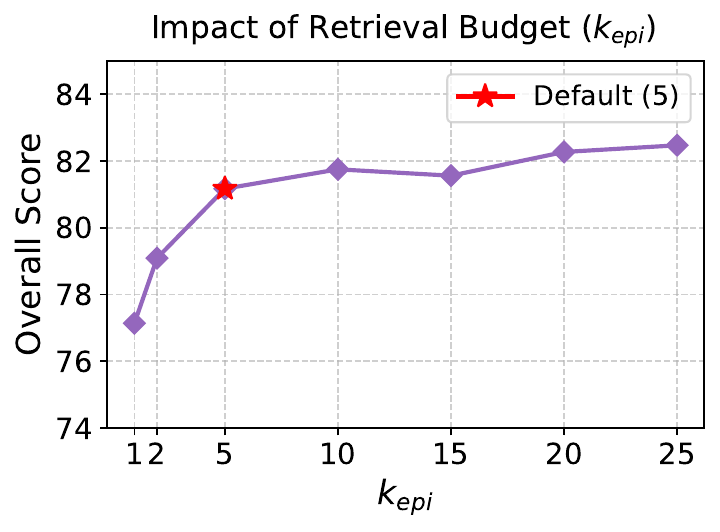}
        \caption{}
        \label{hyper:epi}
    \end{subfigure}
    \caption{Sensitivity of retrieval budgets on LoCoMo (GPT-4.1-mini). (a) Overall score vs. subconscious retrieval budget $k_{\mathrm{sub}}$. (b) Overall score vs. episodic budget $k_{\mathrm{epi}}$ with $k_{\mathrm{sem}}=2k_{\mathrm{epi}}$.}
\end{figure*}
We organize the discussion into two parts: (i) \textbf{consolidation-stage} thresholds, including the recurrence count threshold ($\theta_{\text{count}}$) and the similarity threshold ($\theta_{\text{sim}}$), which determine when interaction clusters are promoted from subconscious memory to higher-level episodic and semantic memories; and (ii) \textbf{retrieval-stage} budgeting, where we cap the number of retrieved items from each memory tier to control context length.
For retrieval, we treat the subconscious and episodic budgets ($k_{\text{sub}}$, $k_{\text{epi}}$) as the only free hyperparameters, and set the semantic budget as a fixed function of the episodic budget, $k_{\text{sem}} = 2k_{\text{epi}}$.

To ensure fair comparison and isolate causal effects, in each experiment we vary only one target hyperparameter and freeze all others to the default LoCoMo configuration.
Unless otherwise stated, we use $\theta_{\text{count}}{=}5$ and $\theta_{\text{sim}}{=}0.7$ for consolidation, and $k_{\text{sub}}{=}10$, $k_{\text{epi}}{=}5$ (thus $k_{\text{sem}}{=}10$) for retrieval.
When sweeping $k_{\text{epi}}$, we update $k_{\text{sem}}$ accordingly via $k_{\text{sem}}=2k_{\text{epi}}$, while keeping all remaining hyperparameters fixed.
All experiments in this section are conducted on LoCoMo using GPT-4.1-mini as the backbone model.

\subsection{Consolidation Hyperparameters}
\label{app:hyper_consolidation}
We study two consolidation-stage thresholds that govern demand-driven memory promotion: the similarity threshold $\theta_{\text{sim}}$, which controls how interaction units are clustered in subconscious memory, and the recurrence threshold $\theta_{\text{count}}$, which controls when a cluster is consolidated into episodic/semantic memories.

\paragraph{Impact of $\theta_{\text{sim}}$.}
As shown in Figure~\ref{hyper:sim}, $\theta_{\text{sim}}$ exhibits a clear peak around the default setting $\theta_{\text{sim}}{=}0.7$.
When $\theta_{\text{sim}}$ is too low, semantically unrelated interactions are merged into the same cluster, reducing topical coherence and making the downstream summarization step noisier.
Conversely, when $\theta_{\text{sim}}$ is too high, related interactions are fragmented across multiple small clusters, weakening recurrence signals and delaying (or preventing) consolidation for genuinely recurring topics.
Overall, the sharp optimum suggests that the best choice on LoCoMo is unambiguous and that RecMem is reasonably robust in the neighborhood of $\theta_{\text{sim}}{=}0.7$.

\paragraph{Impact of $\theta_{\text{count}}$: quality--cost trade-off.}
Figure~\ref{hyper:count} and Figure~\ref{hyper:count_cost} highlight a more explicit effectiveness--efficiency tension for $\theta_{\text{count}}$.
Lower $\theta_{\text{count}}$ triggers consolidation earlier, so each consolidation event typically includes fewer raw interaction units.
This smaller consolidation context can better preserve fine-grained details (less compression pressure during episodic abstraction), but it is also more aggressive and therefore increases construction-time token consumption due to more frequent consolidations.
Accordingly, token cost decreases smoothly as $\theta_{\text{count}}$ increases (Figure~\ref{hyper:count_cost}).

In contrast, the performance curve is not smooth: we observe a clear degradation when increasing $\theta_{\text{count}}$ from $5$ to $6$ (Figure~\ref{hyper:count}), while the corresponding token reduction remains comparatively gradual.
We attribute this drop to two compounding factors at higher thresholds: (i) consolidation becomes overly conservative, leaving some recurring patterns insufficiently represented in episodic/semantic memory at query time; and (ii) once consolidation is finally triggered, the accumulated cluster is larger, which increases summarization difficulty and raises the likelihood that salient details are omitted or poorly organized (even with semantic refinement).
Taken together, these results indicate that $\theta_{\text{count}}{=}5$ is the best operating point on LoCoMo: it retains the accuracy benefits of earlier, detail-preserving consolidation while avoiding unnecessary consolidation overhead, and it prevents the disproportionate quality loss observed at more conservative threshold setting like  $\theta_{\text{count}}{=}6$ (81.1 vs 78.9).

\subsection{Retrieval Hyperparameters}
\label{app:hyper_retrieval}
We next analyze the retrieval-stage budgets that control how much evidence is surfaced from each memory tier at query time.
Recall that we treat the subconscious and episodic budgets ($k_{\text{sub}}$, $k_{\text{epi}}$) as the only free retrieval hyperparameters, and set the semantic budget deterministically as $k_{\text{sem}}=2k_{\text{epi}}$.
Thus, sweeping $k_{\text{epi}}$ implicitly scales the total retrieved memory volume, while sweeping $k_{\text{sub}}$ isolates the contribution of raw, fine-grained interaction evidence.

Figure~\ref{hyper:sub} and Figure~\ref{hyper:epi} show a consistent diminishing-returns trend: increasing retrieval budgets yields substantial gains at small values, but improvements become marginal as budgets grow.
We therefore adopt compact defaults that retain most of the performance benefit while limiting retrieved context length, setting $k_{\text{sub}}{=}10$ and $k_{\text{epi}}{=}5$ (thus $k_{\text{sem}}{=}10$).

\section{Evaluation Datasets}
\label{app:datasets}

This section provides detailed specifications and preprocessing protocols for the two benchmarks used in our experiments: LoCoMo~\cite{locomo} and LongMemEval-S~\cite{longmemeval}.

\subsection{LoCoMo}
LoCoMo (Long-Context Memory) is a benchmark designed to evaluate memory systems in casual, social settings. 
Unlike standard user-agent interactions, the source texts consist of multi-session human-to-human dialogues between two distinct speakers, simulating the natural evolution of a long-term relationship.

\paragraph{Data Statistics.} The dataset consists of 10 independent, human-annotated conversations. Each conversation spans multiple sessions, simulating a relationship that evolves over time.
\begin{itemize}
    \item \textbf{Total Conversations:} 10
    \item \textbf{Average Length:} $\approx$ 16,000 tokens per conversation
    \item \textbf{Total Questions (Used):} 1,540
    \item \textbf{Dialogue Style:} Casual, multi-turn, life-sharing, highly contextual.
\end{itemize}

\paragraph{Task Categories.} The benchmark originally includes five question categories. Following standard protocols established in prior works~\cite{mem0,memoryOS,mirix}, we evaluate on the first four categories and exclude the adversarial set:
\begin{enumerate}
    \item \textbf{Single-hop Retrieval:} Questions requiring the retrieval of a specific fact mentioned in a single past session.
    \item \textbf{Multi-hop Reasoning:} Questions that require synthesizing information distributed across multiple distinct sessions to derive an answer.
    \item \textbf{Temporal Reasoning:} Questions testing the system's ability to understand the sequence of events and relative time expressions.
    \item \textbf{Open-domain Knowledge:} Questions that require combining memory retrieval with external world knowledge.
    \item \textit{Adversarial (Excluded):} Questions designed to trick the model with false premises. We exclude this category as it lacks reliable ground-truth answers for automated evaluation.
\end{enumerate}

\subsection{LongMemEval-S}
LongMemEval-S is a subset of the LongMemEval benchmark, curated to evaluate memory systems in \textit{agentic, task-oriented} interactions with long context windows.

\paragraph{Data Statistics.} Unlike the social nature of LoCoMo, LongMemEval-S features functional interactions where the user seeks specific assistance.
\begin{itemize}
    \item \textbf{Total Conversations:} 500 
    \item \textbf{Average Context Length:} $\approx$ 115k tokens (approx. 30--40 sessions).
    \item \textbf{Total Questions:} 500
    \item \textbf{Dialogue Style:} Task-oriented, high information density.
\end{itemize}

\paragraph{Task Categories.} To assess memory capabilities at a granular level, the benchmark stratifies queries into six distinct types:
\begin{enumerate}
    \item \textbf{Single-session-user:} Evaluates the retrieval of specific details explicitly mentioned by the \textit{user} within the bounds of a single conversation session.
    \item \textbf{Single-session-assistant:} Tests the system's ability to recall information provided by the \textit{assistant} itself within a single session, ensuring consistency in the agent's own history.
    \item \textbf{Single-session-preference:} Assesses whether the model can effectively apply retrieved user information to generate personalized, context-aware responses.
    \item \textbf{Multi-session:} Requires the aggregation of disjoint pieces of information scattered across two or more sessions to derive a complete answer.
    \item \textbf{Knowledge-update:} Probes the system's capacity to track dynamic changes in the user's life state and supersede outdated information with new updates.
    \item \textbf{Temporal-reasoning:} Demands chronological deduction by synthesizing both the session metadata (timestamps) and explicit time expressions found in the text.
\end{enumerate}
\section{Experiment Details}
\label{app:experiment}

\subsection{Baseline Configurations}
\label{app:baseline}
To ensure fair and reliable comparisons, we configure each baseline to faithfully reflect its original design choices, rather than enforcing a unified ingestion or prompting pipeline. Below, we describe the implementation and prompting decisions used in our experiments in details.

To enable a fair comparison of computational costs, we instrumented all baseline codebases with unified token-tracking logic while leaving their core memory components intact. For the LoCoMo benchmark, all memory-system baselines considered in this work provide official implementations. We therefore reuse their original prompts and evaluation code without modification.

For the LongMemEval-S benchmark, where standardized reference implementations are not available, we implement the evaluation pipeline while preserving each method's ingestion strategy as used in its LoCoMo setup. Concretely, we adopt: (i) A-Mem's per-message ingestion, (ii) Mem0's dual-speaker ingestion with two messages per turn, and (iii) MemoryOS's ingestion based on user--assistant QA pairs. We make this choice to respect the baselines' intended memory abstractions; forcing all methods to share RecMem's ingestion logic would conflate design differences and bias the comparison.

For A-Mem and MemoryOS, they both have two official codebases and we adopt the ones used in their paper~\cite{a-mem,memoryOS} to ensure reproduction of the reported setting. For Mem0, we use its local-deployment version to enable token-consumption tracking. We also disable graph construction, as Mem0 reports that its graph variant can lead to a performance drop~\cite{mem0}.

For the RAG-2048 baseline, we adopt a conservative chunking strategy that preserves message integrity: we never split a message across two chunks. Messages are accumulated sequentially until the chunk reaches the 2048-token budget. If adding the next message would exceed this limit, we still include the entire message (rather than truncating it) to preserve semantic completeness, and then start a new chunk from the subsequent message.


\subsection{Evaluation Prompt Consistency}

To ensure a fair and standardized comparison, we strictly enforce prompt consistency across all evaluated methods. For any given dataset, the exact same evaluation prompt is employed for the LLM judge across all baselines and RecMem, ensuring that performance differences originate solely from the memory systems' capabilities rather than variations in the evaluation criteria. Specifically, our prompt sources are as follows: 

\paragraph{LongMemEval-S}: We adopt the official evaluation prompt provided by the benchmark authors~\cite{longmemeval} without modification. 

\paragraph{LoCoMo}: We follow the evaluation protocol established in previous work~\cite{mem0}, which adapts the evaluation prompt elements originally designed by MemGPT~\cite{memgpt}. 

\subsection{Answer Prompt Consistency}
\label{app:anwser_prompt_consistency}
For LoCoMo, we use each baseline's original answering prompt. For LongMemEval-S, we use the main body of RecMem's answering prompt as a shared answer template across methods, so that performance differences primarily reflect the underlying memory mechanisms rather than prompt engineering. For the \textbf{Full-Context} and \textbf{RAG-2048} baselines, we also use the same answering prompt as RecMem for consistency.

Because \textbf{RecMem} and \textbf{MemoryOS} both adopt multi-module memory architectures, their answering prompts include a short module description that clarifies the roles of different memory sources. For MemoryOS, we retain the prompt format used in its LoCoMo implementation. For RecMem, we include a brief module-role description to prevent the answer agent from double-counting overlapping evidence retrieved from different modules. Baselines with a single memory source do not require such clarification and therefore use only the shared main prompt body.

\subsection{Discussion on F1 Score}
\label{app:F1}

While the F1 score is one of the standard metrics in prior works~\cite{a-mem,memoryOS,mem0}, measuring token-level exact matching, we observed it to be unreliable for evaluating long-context memory systems where semantic correctness is paramount. 
F1 score penalizes correct answers that differ in phrasing from the ground truth. For instance, if the ground truth is ``16 March, 2023'', and the model generates ``Gina opened her online clothing store on 2023-03-16'', the F1 score approaches 0 despite the answer being factually correct. Consequently, we prioritize LLM-as-Judge in our main analysis.

For LoCoMo, many prior evaluations treat F1 as a primary metric and enforce strict output-length constraints (e.g., ``the answer should be less than 5 words'') to optimize token overlap. To maintain comparability with these reporting conventions, we retain such length constraints when evaluating on this dataset. For transparency, we additionally report the resulting F1 scores in Table~\ref{tab:locomo-f1}. 
In contrast, for LongMemEval-S, since all methods utilize a shared prompt body, we remove these artificial constraints to avoid penalizing valid, grounded answers that may exceed rigid word counts.

\begin{table}[ht]
\centering
\small
\setlength{\tabcolsep}{4pt}
\begin{tabular}{p{2.6cm}|cc}
\hline
\textbf{Metrics}
& \textbf{F1 Score}
& \textbf{Judge Score } \\
\hline
\multicolumn{3}{l}{\textsc{gpt-4o-mini}}\\
\hline
Full Context    & 0.423 & \textbf{76.4} \\
Naive Rag       & 0.309 & 49.7 \\
MemoryOS        & \textbf{0.456} & 63.6 \\
Mem0            & \underline{0.429} & 58.6 \\
AMem            & 0.364 & 60.8 \\
RecMem(Ours)     & 0.385 & \underline{72.47} \\
\hline
\multicolumn{3}{l}{\textsc{gpt-4.1-mini}}\\
\hline
Full Context    & \textbf{0.507} & \textbf{84.8} \\
Naive Rag       & 0.337 & 54.4 \\
MemoryOS        & 0.445 & 67.6 \\
Mem0            & 0.428 & 62.9 \\
AMem            & 0.440 & 69.7 \\
RecMem(Ours)     & \underline{0.469} & \underline{81.1} \\
\hline
\end{tabular}
\caption{F1 score and llm judge score on LoCoMo.}
\label{tab:locomo-f1}
\end{table}
\subsection{Retrieval Top-K}

For \textbf{RAG-2048}, we set the retrieval top-$K$ to $3$ on LoCoMo, reflecting its relatively short conversation length, and to $5$ on LongMemEval-S, where conversations are longer and often require aggregating evidence across more chunks. As shown by the query-token statistics in Tables~\ref{tab:longmemeval-main} and~\ref{tab:locomo-main}, these settings allow the RAG baseline to retrieve a comparable amount of information to other methods under similar query-time budgets.

For all other baselines, we keep their retrieval budgets consistent across LoCoMo and LongMemEval-S, following their default design choices. Concretely, \textbf{A-Mem} retrieves $10$ memory notes. \textbf{Mem0} retrieves $60$ memory facts in total. \textbf{MemoryOS} retrieves all memories from short-term memory, together with $10$ memories from mid-term memory, $5$ memories from long-term memory, as well as its qualified assistant knowledge and user knowledge components.

A special case is \textbf{Mem0} on LoCoMo: since LoCoMo includes dual-speaker question types, Mem0 retrieves $30$ facts per speaker (i.e., $60$ total) to balance coverage across user and assistant perspectives. In contrast, LongMemEval-S is dominated by user-centric questions, with relatively few assistant-centric queries. Therefore, while keeping the total budget fixed at $60$, we allocate $45$ retrieved facts to the user side and $15$ to the assistant side, which better reflects Mem0's intended strengths under the LongMemEval-S query distribution.
\label{app:ablation}

\section{LLM Prompts}
\label{app:prompts}

This appendix reports the primary prompts used in \textsc{RecMem}, including
(i) episodic memory generation,
(ii) episodic memory merging,
(iii) semantic memory generation, and
(iv) the final answer prompt.
To improve readability and facilitate reproduction, we present each prompt in figures
instead of inline text.
Each memory-related prompt follows a consistent structure with three components:
(a) a role and goal specification,
(b) detailed instructions, and
(c) explicit output-format constraints.

\begin{figure*}
    \centering
    \includegraphics[width=\linewidth]{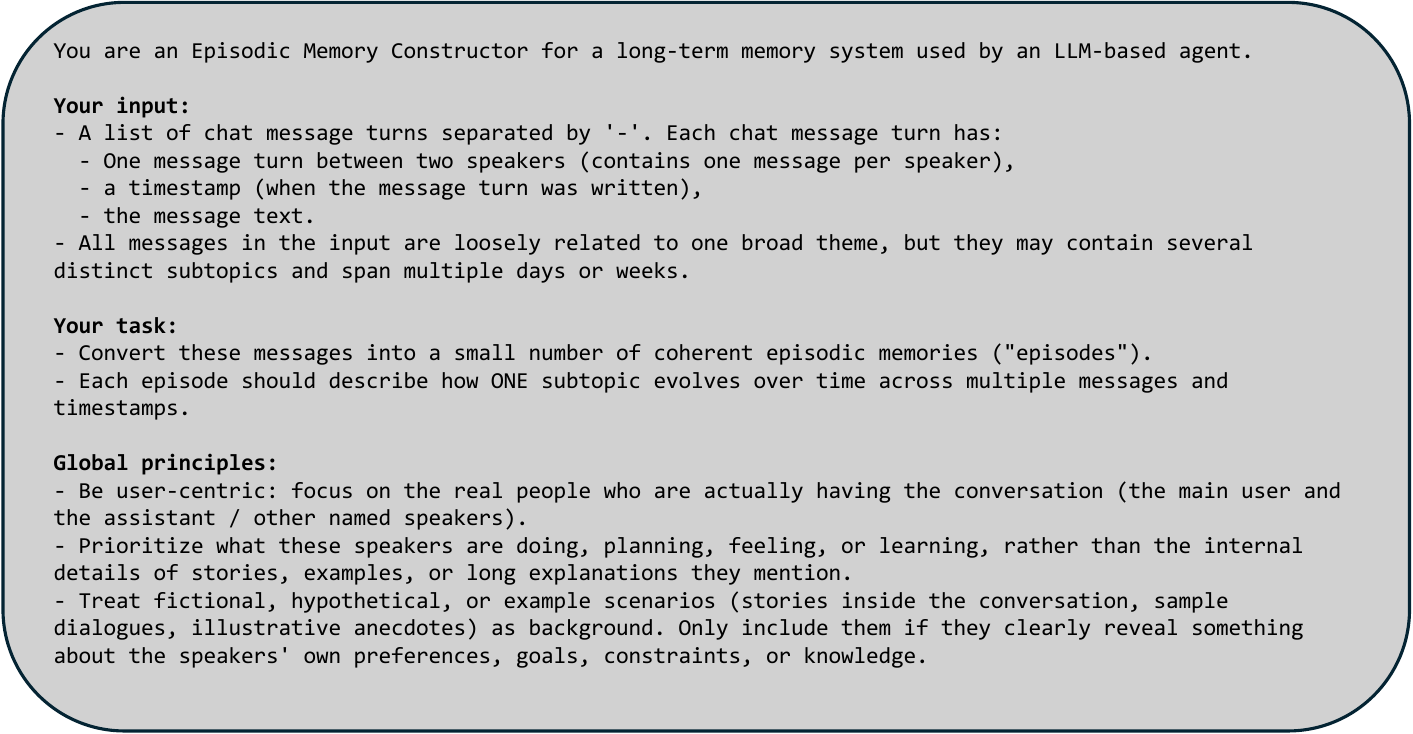}
    \caption{Episodic Memory Generation Role Description}
    \label{prompt:epi_role}
\end{figure*}

\begin{figure*}
    \centering
    \includegraphics[width=\linewidth]{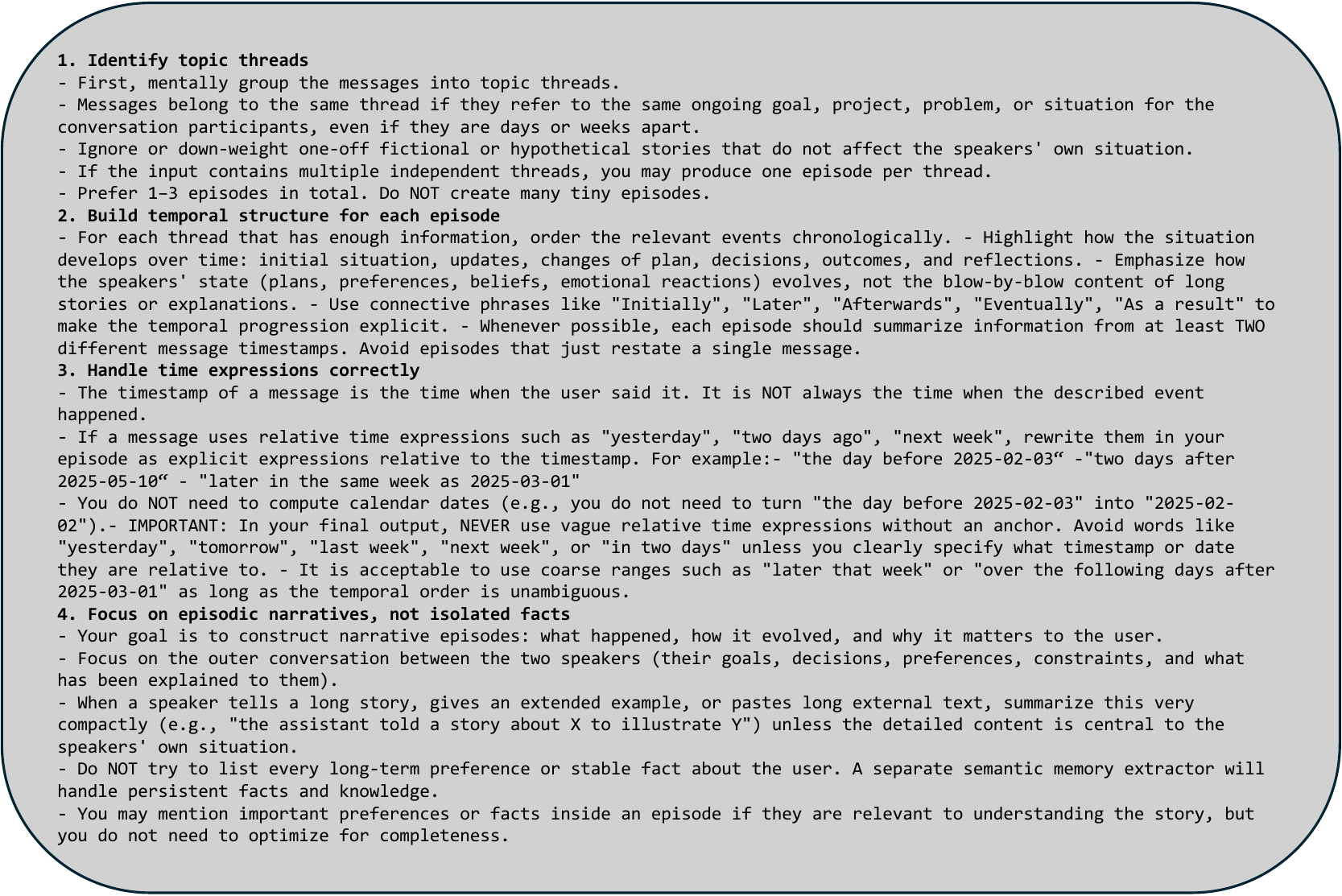}
    \caption{Episodic Memory Generation Instruction}
    \label{prompt:epi_instruction}
\end{figure*}

\begin{figure*}
    \centering
    \includegraphics[width=0.9\linewidth]{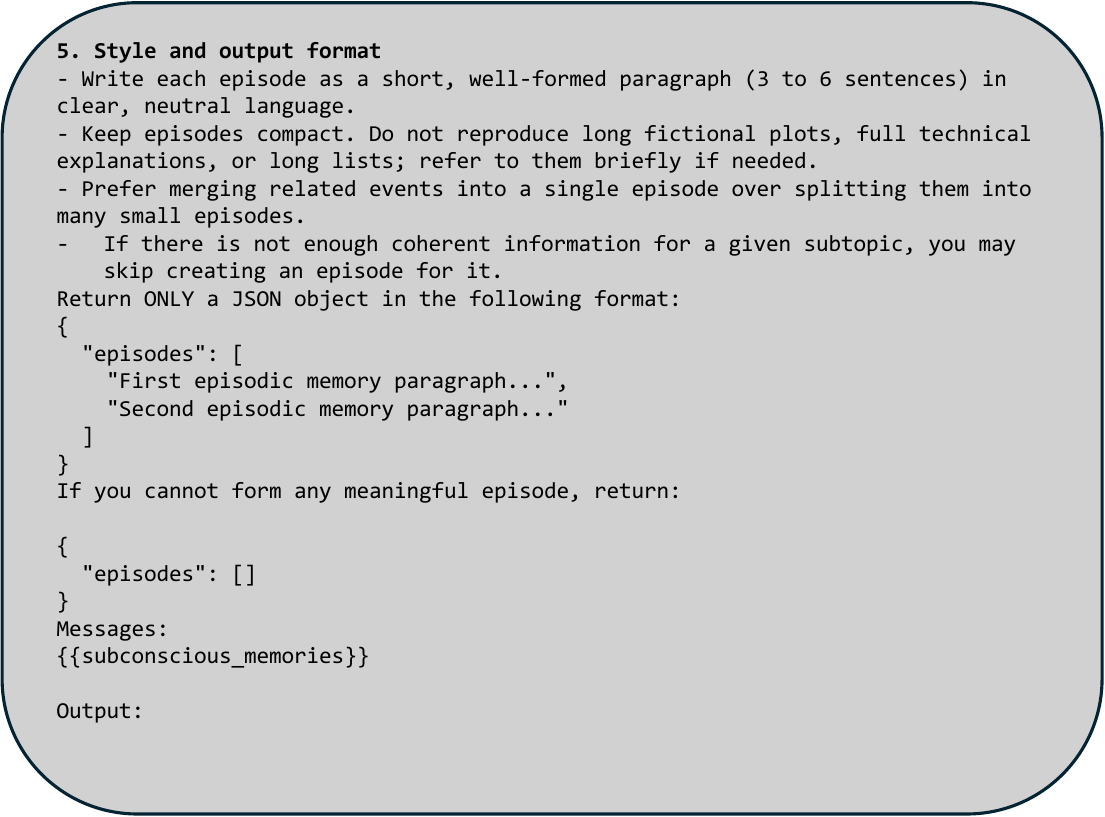}
    \caption{Episodic Memory Output Format}
    \label{prompt:epi_output}
\end{figure*}

\paragraph{Episodic memory generation prompt.}
Figures~\ref{prompt:epi_role}--\ref{prompt:epi_output} show the role/goal description,
instructions, and required output format for episodic memory generation.

\begin{figure*}
    \centering
    \includegraphics[width=0.9\linewidth]{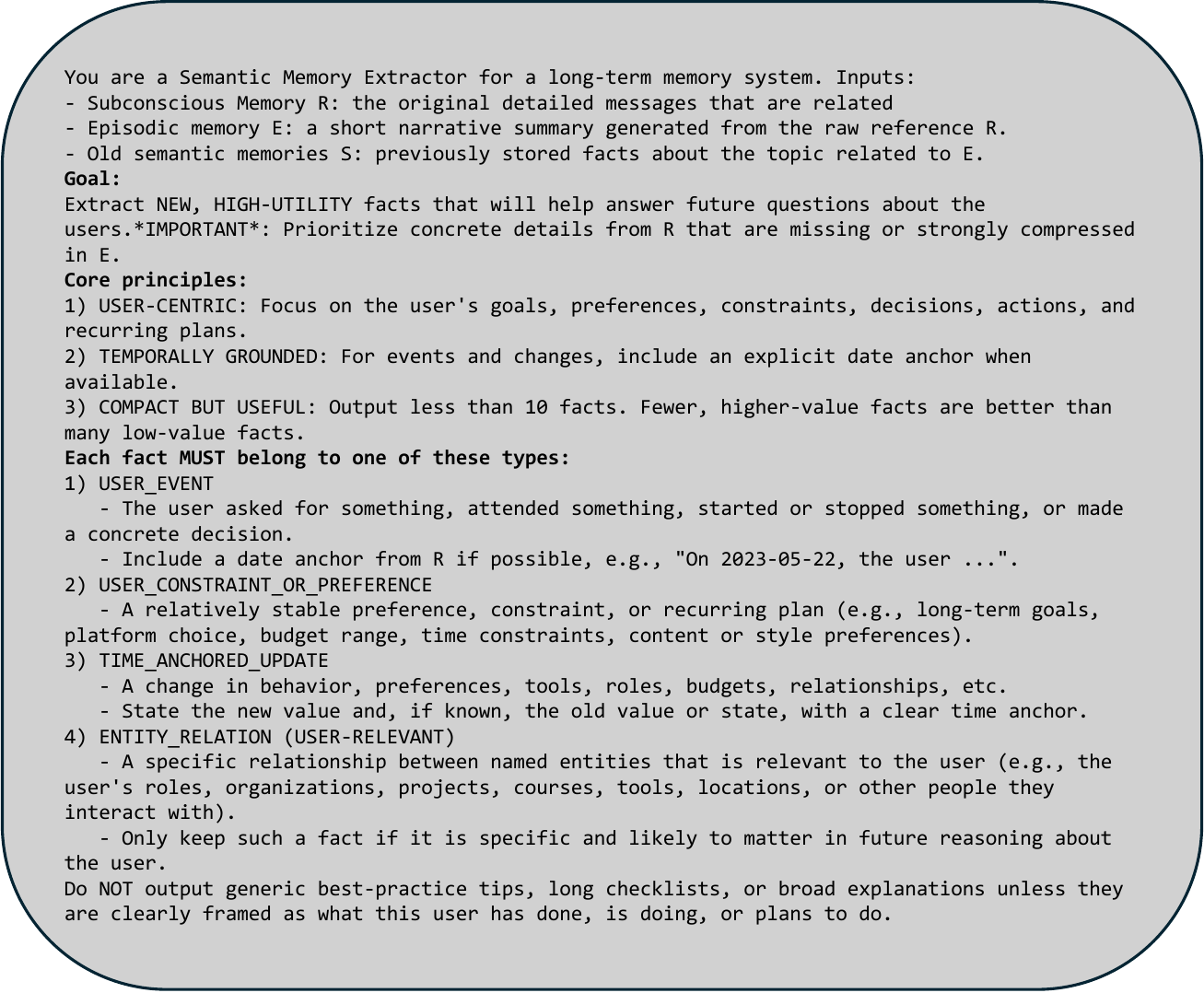}
    \caption{Semantic Memory Generation Role Description}
    \label{prompt:sem_role}
\end{figure*}

\begin{figure*}
    \centering
    \includegraphics[width=0.9\linewidth]{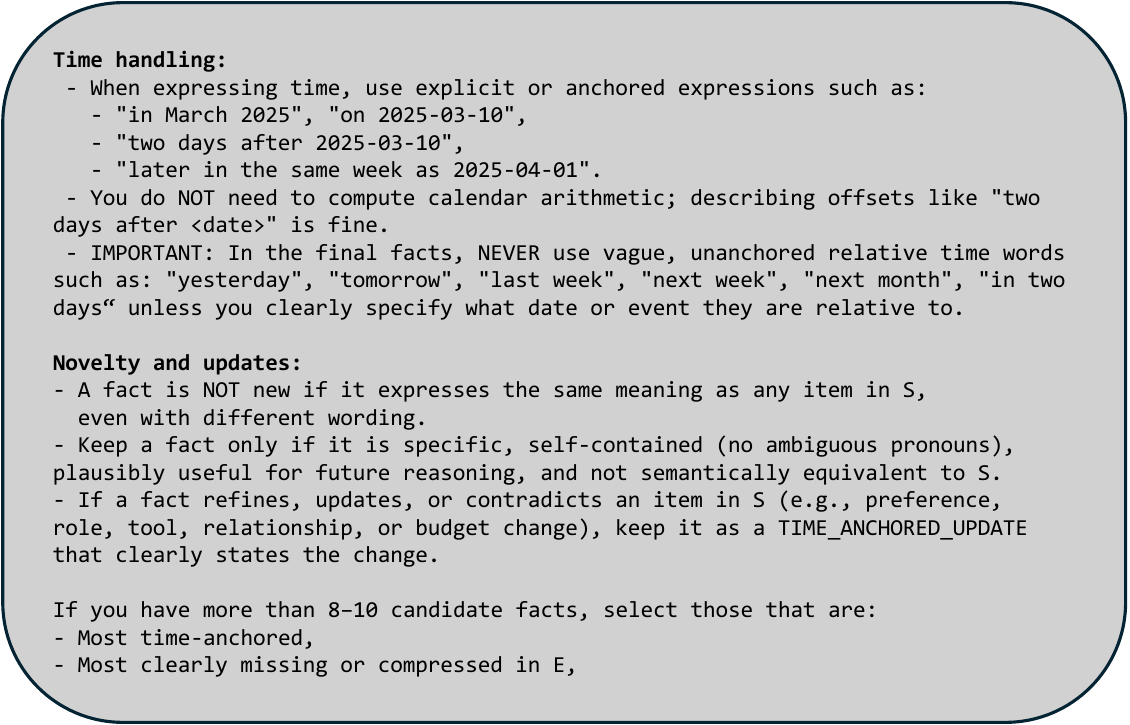}
    \caption{Semantic Memory Generation Instruction}
    \label{prompt:sem_instruction}
\end{figure*}

\begin{figure*}
    \centering
    \includegraphics[width=0.6\linewidth]{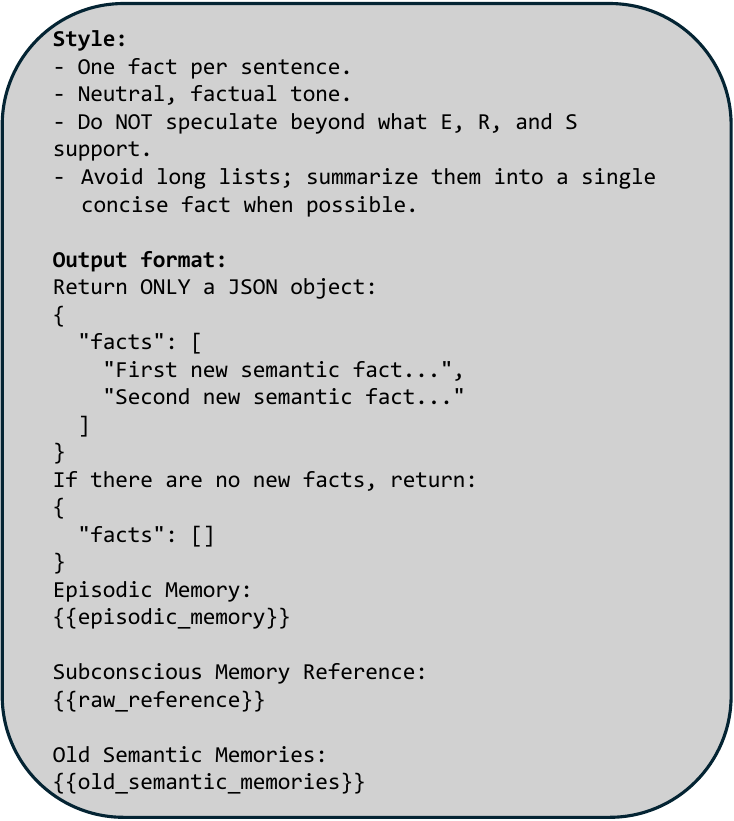}
    \caption{Semantic Memory Output Format}
    \label{prompt:sem_output}
\end{figure*}

\paragraph{Semantic memory generation prompt.}
Figures~\ref{prompt:sem_role}--\ref{prompt:sem_output} present the corresponding components
for semantic memory generation.

\begin{figure*}
    \centering
    \includegraphics[width=0.8\linewidth]{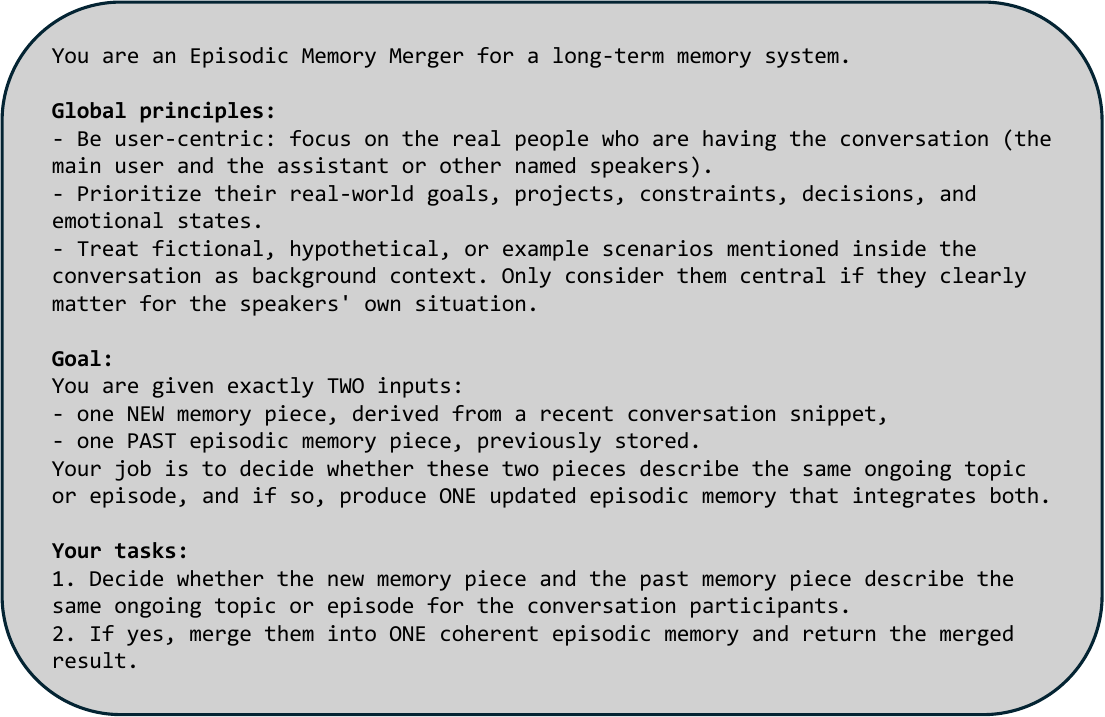}
    \caption{Episodic Merging Role Description}
    \label{prompt:merge_role}
\end{figure*}

\begin{figure*}
    \centering
    \includegraphics[width=\linewidth]{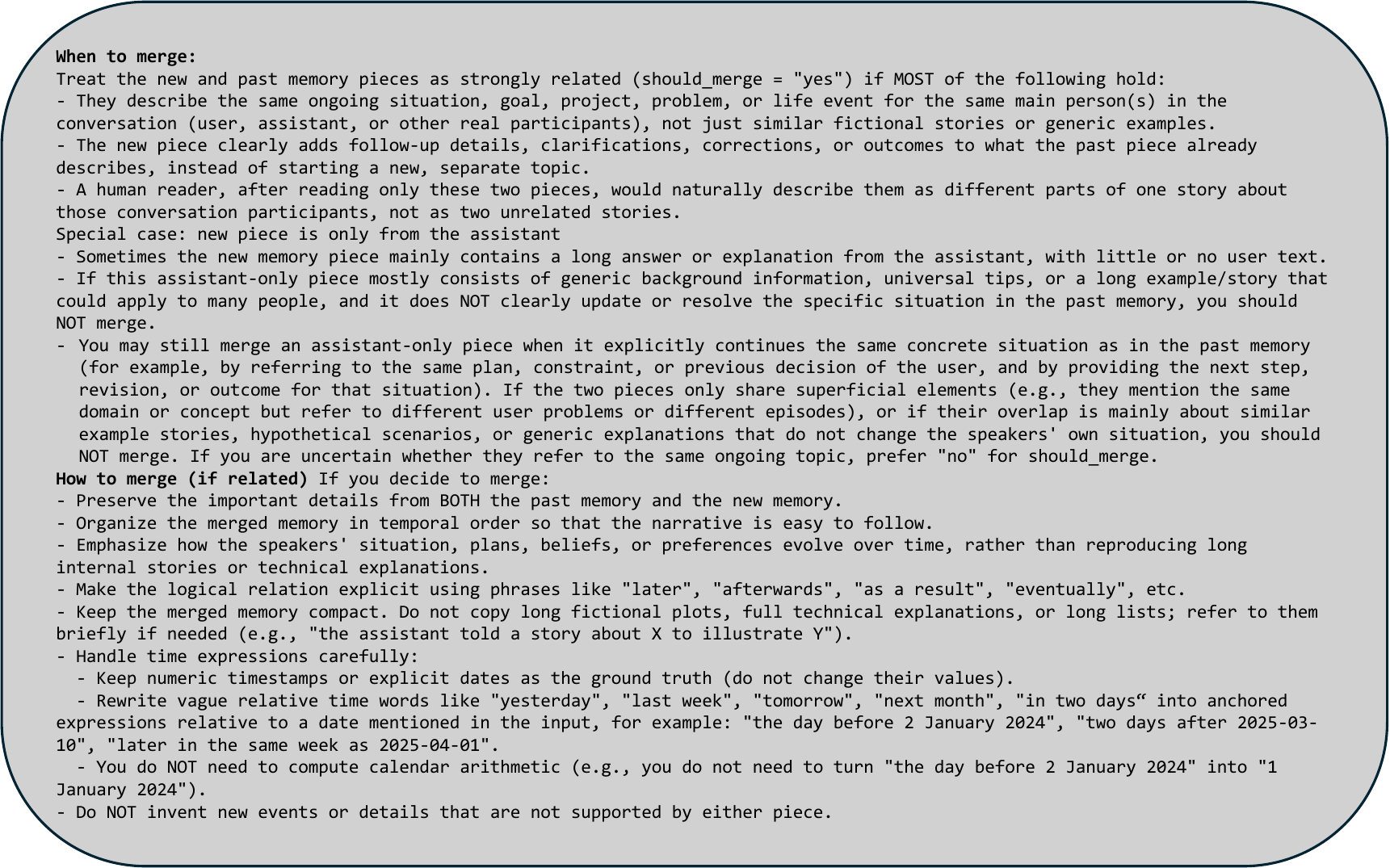}
    \caption{Episodic Merging Instruction}
    \label{prompt:merge_instruction}
\end{figure*}

\begin{figure*}
    \centering
    \includegraphics[width=0.7\linewidth]{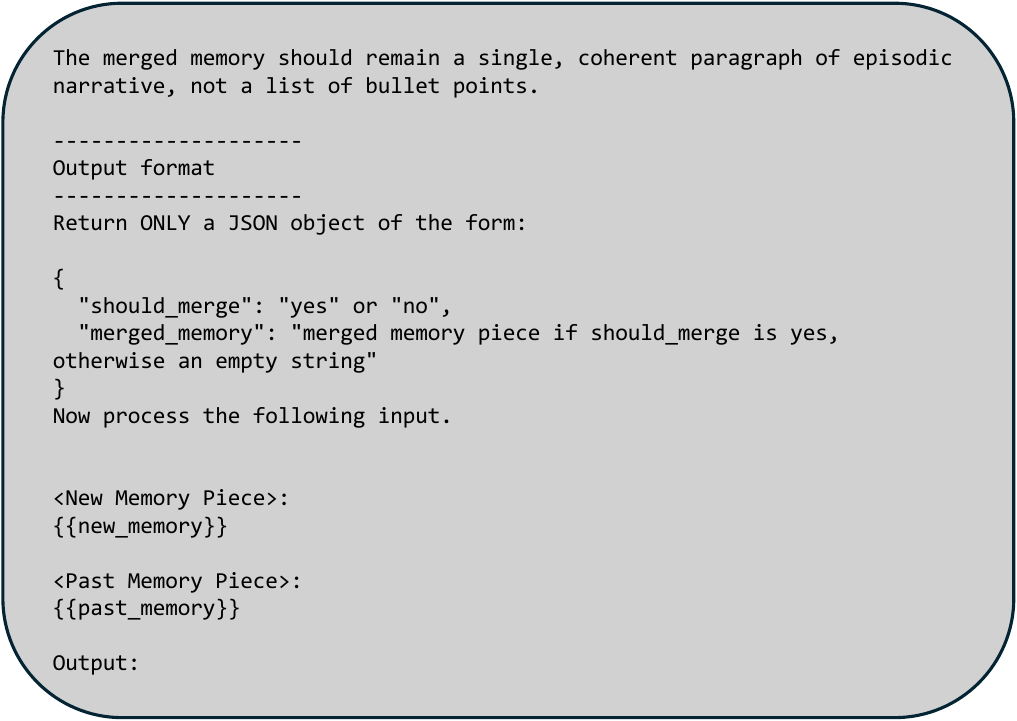}
    \caption{Episodic Merging Output Format}
    \label{prompt:merge_output}
\end{figure*}

\paragraph{Episodic memory merging prompt.}
Figures~\ref{prompt:merge_role}--\ref{prompt:merge_output} provide the prompt used to merge
newly consolidated content into existing episodic memories.

\begin{figure*}
    \centering
    \includegraphics[width=\linewidth]{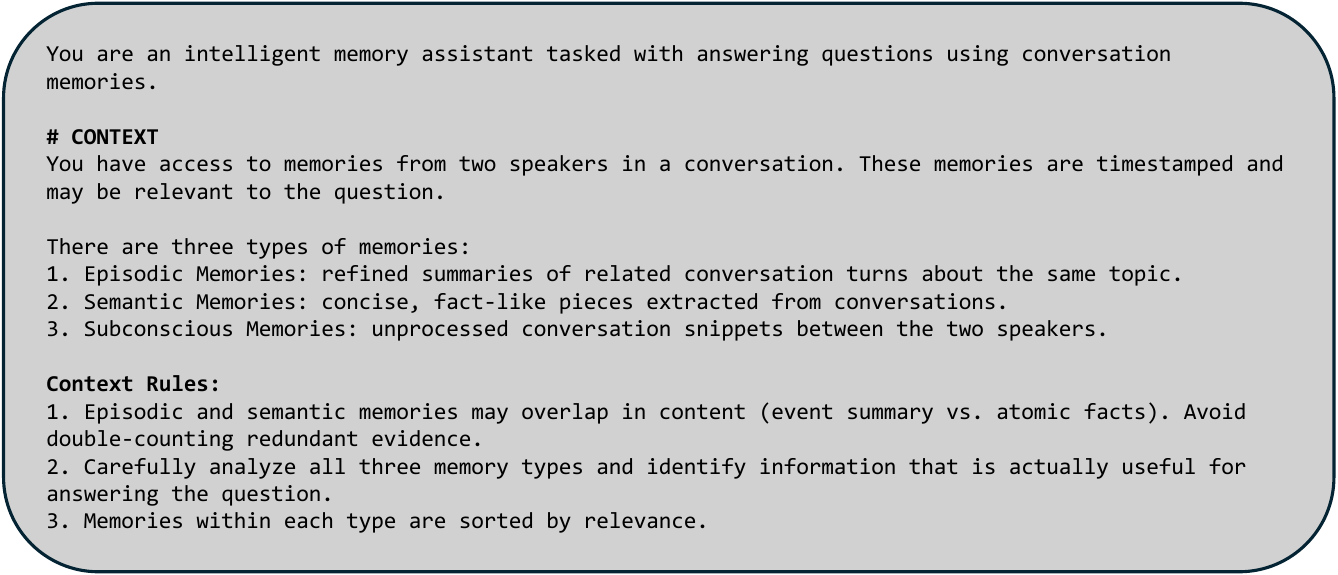}
    \caption{Answering Role Description}
    \label{prompt:answer_role}
\end{figure*}

\begin{figure*}
    \centering
    \includegraphics[width=\linewidth]{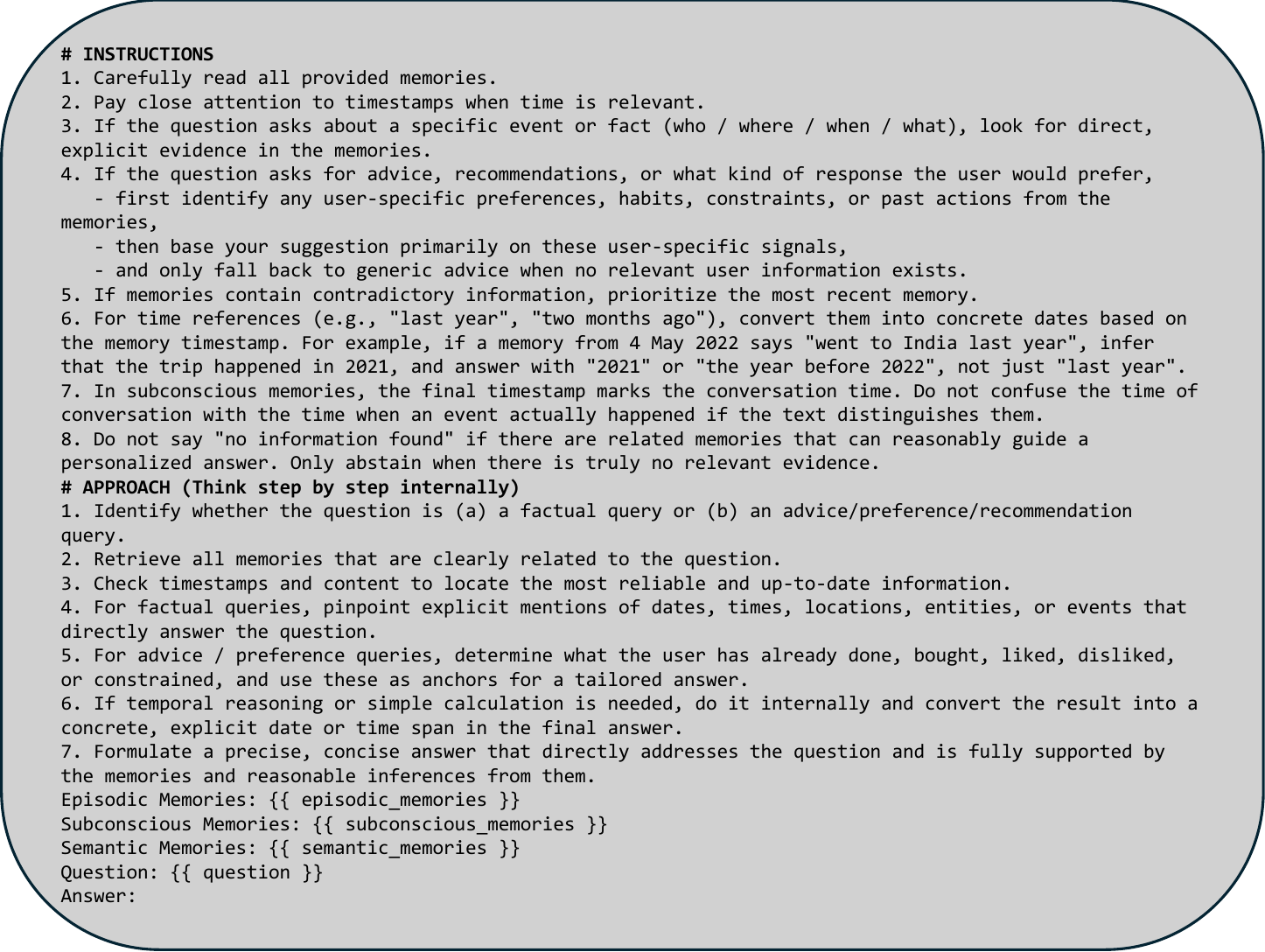}
    \caption{Answering Instruction}
    \label{prompt:answer_instruction}
\end{figure*}

\paragraph{Answer prompt.}
Figures~\ref{prompt:answer_role} and~\ref{prompt:answer_instruction} report the role/goal
and instruction components of the answer prompt used during evaluation.

\section{Licenses and Terms of Use}

\paragraph{Licenses.}
We use publicly released benchmarks under their original licenses: LoCoMo (CC BY-NC 4.0) and LongMemEval-S (MIT License). We do not redistribute these datasets; instead, we refer readers to their official releases. For baselines, we use publicly available implementations under the licenses stated in their official repositories (Mem0: Apache License 2.0; A-Mem: MIT License; MemoryOS: Apache License 2.0). We do not repackage or redistribute third-party artifacts beyond what is permitted by their original licenses.

\paragraph{Terms of Use.}
LoCoMo and LongMemEval-S were released as research benchmarks for evaluating conversational assistants. We use them strictly in the intended offline evaluation setting, following the benchmark protocols. We do not redistribute the datasets and only report aggregated results, consistent with their stated licenses and access conditions. RecMem is evaluated on these benchmarks, but its core mechanisms are applicable to a broader class of long-running conversational agent settings and can be integrated into practical systems. Real-world deployment should be adapted to the target workflow and comply with applicable data licenses/terms and usage conditions.

\end{document}